\documentclass[10pt,twocolumn,letterpaper]{article}

\usepackage{cvpr}
\usepackage{times}
\usepackage{epsfig}
\usepackage{graphicx}
\usepackage{amsmath}
\usepackage{amssymb}
\usepackage{array}

\usepackage{subcaption}

\usepackage[pagebackref=true,breaklinks=true,letterpaper=true,colorlinks,bookmarks=false]{hyperref}

\cvprfinalcopy 

\def\cvprPaperID{3805} 

\ifcvprfinal\fi
\usepackage{todo}
\usepackage{xcolor}
\usepackage[normalem]{ulem}

\newcommand\comment[1]{}
\newcommand{\figsz}{0.47in}
\newcolumntype{V}{>{\centering\arraybackslash} m{\figsz} }
\newcommand{\showtexframe}[1]{\epsfig{file=#1, width = \figsz}}
\newcommand{\showtexture}[1]{
\vspace{0.1cm} \showtexframe{#10.jpeg} \showtexframe{#11.jpeg} \showtexframe{#12.jpeg} \showtexframe{#13.jpeg} \showtexframe{#14.jpeg} \showtexframe{#15.jpeg} \showtexframe{#16.jpeg} \showtexframe{#17.jpeg} \showtexframe{#18.jpeg} \showtexframe{#19.jpeg}
}
\newcommand{\showtextureshort}[1]{
\vspace{0.1cm} \showtexframe{#10.jpeg} \showtexframe{#11.jpeg} \showtexframe{#12.jpeg} \showtexframe{#13.jpeg} \showtexframe{#14.jpeg}
}
\usepackage{csvsimple}
\usepackage{rotating}
\usepackage{subcaption}
\usepackage{lscape}

\graphicspath{{figures/}}

\title{Two-Stream Convolutional Networks for Dynamic Texture Synthesis}

\author{
  Matthew Tesfaldet\quad Marcus A. Brubaker \\
  Department of Electrical Engineering and Computer Science \\
  York University, Toronto \\
  \small\texttt{\{mtesfald,mab\}@eecs.yorku.ca} \\
  \and
  Konstantinos G. Derpanis \\
  Department of Computer Science \\
  Ryerson University, Toronto \\
  \small\texttt{kosta@scs.ryerson.ca}
}

\begin{document}

\maketitle
\thispagestyle{empty} 

\begin{abstract}
We introduce a two-stream model for dynamic texture synthesis.
Our model is based on pre-trained convolutional networks (ConvNets)
that target two independent tasks: (i) object recognition, and (ii)
optical flow prediction.
Given an input dynamic texture, statistics
of filter responses from the object recognition ConvNet
encapsulate the per-frame appearance of the input texture, while
statistics of filter responses from the optical flow ConvNet model
its dynamics.
To generate a novel texture, a randomly initialized input sequence is optimized
to match the feature statistics from each
stream of an example texture.  
Inspired by recent work on image style transfer and enabled by the
two-stream model, we also apply the synthesis approach to combine the
texture appearance from one texture with the dynamics of another 
to generate entirely novel dynamic textures.
We show that our approach generates novel, high quality samples 
that match both the framewise appearance and temporal evolution
of input texture.
Finally, we quantitatively evaluate our texture synthesis approach with a thorough user study.
\end{abstract}

\section{Introduction}

\begin{figure}[t]
\begin{center}
	\epsfig{file=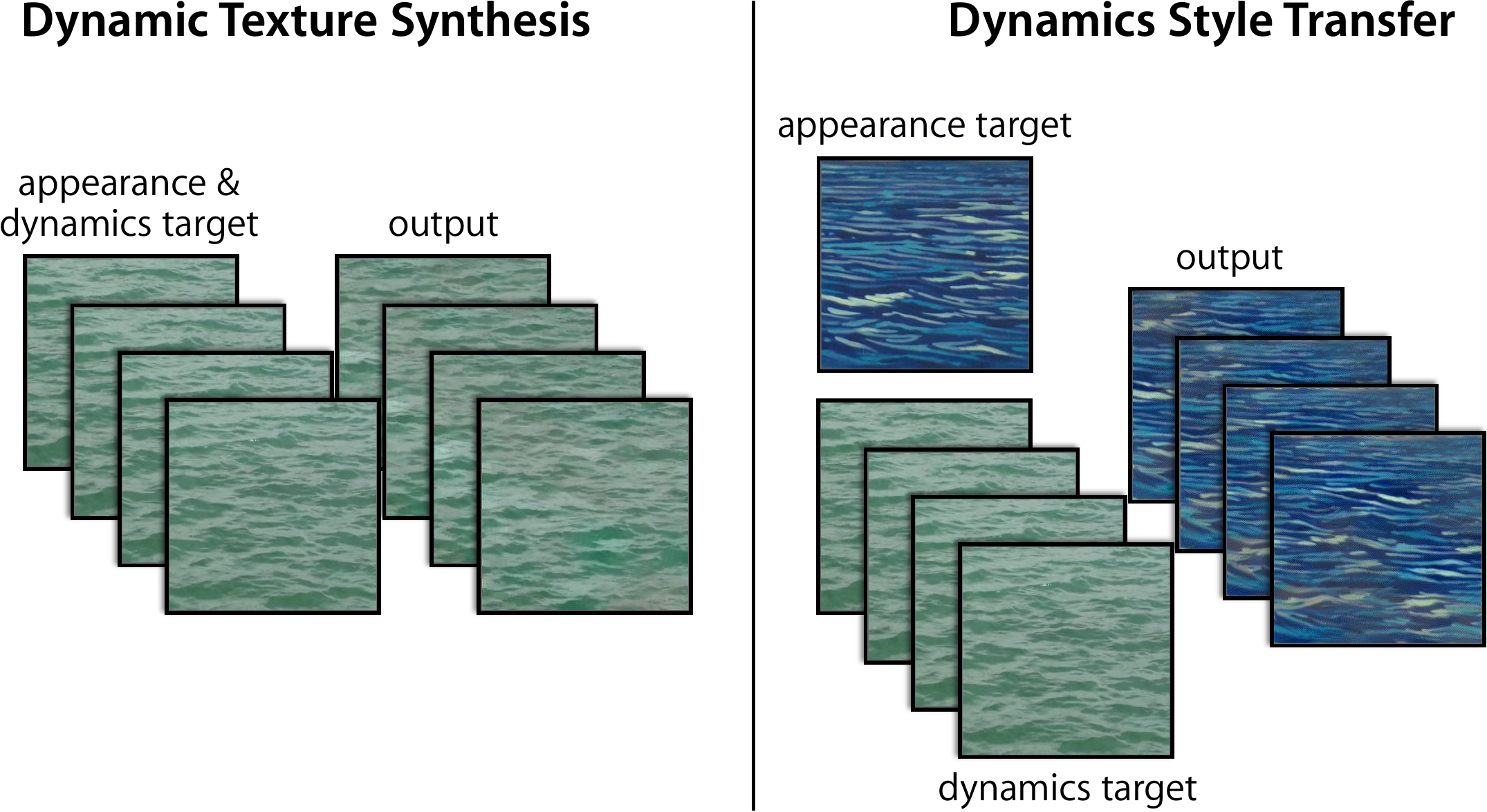, width = 0.475\textwidth}\\
	\caption{Dynamic texture synthesis. (left) Given an input dynamic texture as the target, our two-stream model is able to synthesize a novel dynamic texture that preserves the target's appearance and dynamics characteristics. (right) Our two-stream approach enables synthesis that combines the texture appearance from one target with the dynamics from another, resulting in a composition of the two.}
	\vspace{-0.65cm}
	\label{fig:teaser}
\end{center}
\end{figure}

Many common temporal visual patterns are naturally described by
the ensemble of appearance and dynamics (\ie, temporal pattern
variation) of their constituent elements. 
Examples of such patterns include fire, fluttering vegetation, and wavy
water.
Understanding and characterizing these temporal patterns has
long been a problem of interest in human perception, computer
vision, and computer graphics.
These patterns have been previously studied under a variety of names, including
turbulent-flow motion \cite{heeger1986}, temporal textures
\cite{nelson1992}, time-varying textures \cite{bar-joseph2001},
dynamic textures \cite{doretto2003}, textured motion \cite{wang2003} 
and spacetime textures \cite{derpanis2012spacetime}.
Here, we adopt the term ``dynamic texture''.
In this work, we propose a factored analysis of dynamic textures in
terms of appearance and temporal dynamics.
This factorization is then used to enable dynamic texture synthesis
which, based on example texture inputs, generates a novel dynamic
texture instance.
It also enables a novel form of style transfer where the 
target appearance and dynamics can be taken from different sources
as shown in Fig.\ \ref{fig:teaser}.

Our model is constructed from two convolutional networks
(ConvNets), an appearance stream and a dynamics stream,
which have been pre-trained for object recognition
and optical flow prediction, respectively.
Similar to previous work on spatial textures
\cite{gatys2015,heeger1995pyramid,portilla2000parametric}, we
summarize an input dynamic texture in terms of a set of
spatiotemporal statistics of filter outputs from each stream.
The appearance stream models the per frame appearance of
the input texture, while the dynamics stream models its
temporal dynamics.
The synthesis process consists of optimizing a randomly initialized noise pattern such that its spatiotemporal statistics from
each stream match those of the input texture.
The architecture is inspired by insights from human perception and 
neuroscience.
In particular, psychophysical studies \cite{cutting1982} show that
humans are able to perceive the structure of a dynamic texture even
in the absence of appearance cues, suggesting that the two streams
are effectively independent.
Similarly, the two-stream hypothesis \cite{goodale1992} models the 
human visual cortex in terms of two pathways, the ventral stream
(involved with object recognition) and the
dorsal stream (involved with motion processing).

In this paper, our two-stream analysis of
dynamic textures is applied to texture synthesis.
We consider a range of dynamic textures and show that 
our approach generates novel, high quality samples that match
both the frame-wise appearance and temporal evolution of an input
example.
Further, the factorization of appearance and dynamics enables a 
novel form of style transfer, where dynamics of one texture are 
combined with the appearance of a different one,
\cf\ \cite{gatys2016image}.
This can even be done using a single image as an appearance
target, which allows static images to be animated.
Finally, we validate the perceived realism of our generated textures
through an extensive user study.

\section{Related work}
There are two general approaches that have dominated the texture
synthesis literature: non-parametric sampling approaches that
synthesize a texture by sampling pixels of a given source texture
\cite{efros1999,kwatra2003graphcut,schodl2000,wei2000}, and 
statistical parametric models.
As our approach is an instance of a parametric model, here we focus
on these approaches.

The statistical characterization of visual textures was introduced
in the seminal work of Julesz \cite{julesz1962}.
He conjectured that particular statistics of pixel intensities
were sufficient to partition spatial textures into metameric (\ie,
perceptually indistinguishable) classes.
Later work leveraged this notion for texture synthesis
\cite{heeger1995pyramid,portilla2000parametric}.
In particular, inspired by models of the early stages of visual 
processing, statistics of (handcrafted) multi-scale oriented filter 
responses were used to optimize an initial noise pattern 
to match the filter response statistics of an input texture.
More recently, Gatys et al.\ \cite{gatys2015} demonstrated
impressive results by replacing the linear filter bank with a
ConvNet that, in effect, served as a proxy for the ventral visual
processing stream.
Textures are modelled in terms of the correlations between filter 
responses within several layers of the network.
In subsequent work, this texture model was used in image style
transfer \cite{gatys2016image}, where the style of one image was
combined with the image content of another to produce a new image.
Ruder et al.\ \cite{ruder2016} extended this model to video by using
optical flow to enforce temporal consistency of the
resulting imagery.

Variants of linear autoregressive models have been studied
\cite{szummer1996,doretto2003} that jointly model appearance and
dynamics of the spatiotemporal pattern.
More recent work has considered ConvNets as a basis for modelling 
dynamic textures.
Xie et al.\ \cite{xie2017synthesizing} proposed a spatiotemporal
generative model where each dynamic texture is modelled as a random
field defined by multiscale, spatiotemporal ConvNet filter responses
and dynamic textures are realized by sampling the model.
Unlike our current work, which assumes pretrained fixed networks,
this approach requires the ConvNet weights to be trained using the
input texture prior to synthesis.

A recent preprint \cite{funke2017} described preliminary 
results extending the framework of Gatys et al.\ \cite{gatys2015} 
to model and  synthesize dynamic textures by computing a Gram 
matrix of filter activations over a small temporal window.
In contrast, our two stream filtering architecture is more 
expressive as our dynamics stream is specifically tuned to 
spatiotemporal dynamics.
Moreover, as will be demonstrated, the factorization
in terms of appearance and dynamics enables a novel form of
style  transfer, where the dynamics of one pattern are 
transferred to the appearance of another to generate an
entirely new dynamic texture.
To the best of our knowledge, we are the first to demonstrate 
this form of style transfer.

The recovery of optical flow from temporal imagery has long been
studied in computer vision.
Traditionally, it has been addressed by handcrafted approaches
\eg, \cite{horn1981,lucas1981,revaud2015epicflow}.
Recently, ConvNet approaches \cite{dosovitskiy2015,ranjan2017,ilg2017,yu2016}
have been demonstrated as viable alternatives.
Most closely related to our approach are energy models of visual
motion \cite{adelson1985spatiotemporal,heeger1988,simoncelli1998,nishimoto2011,derpanis2012spacetime,konda2014}
that have been motivated and studied in a variety of contexts,
including computer vision, visual neuroscience, and visual
psychology.
Given an input image sequence, these models consist of an
alternating sequence of linear and non-linear operations that yield
a distributed representation (\ie,  implicitly coded) of pixelwise
optical flow.
Here, an energy model motivates the
representation of observed dynamics which is then encoded
as a ConvNet.

\section{Technical approach}
\begin{figure}[t]
\begin{center}
    \epsfig{file=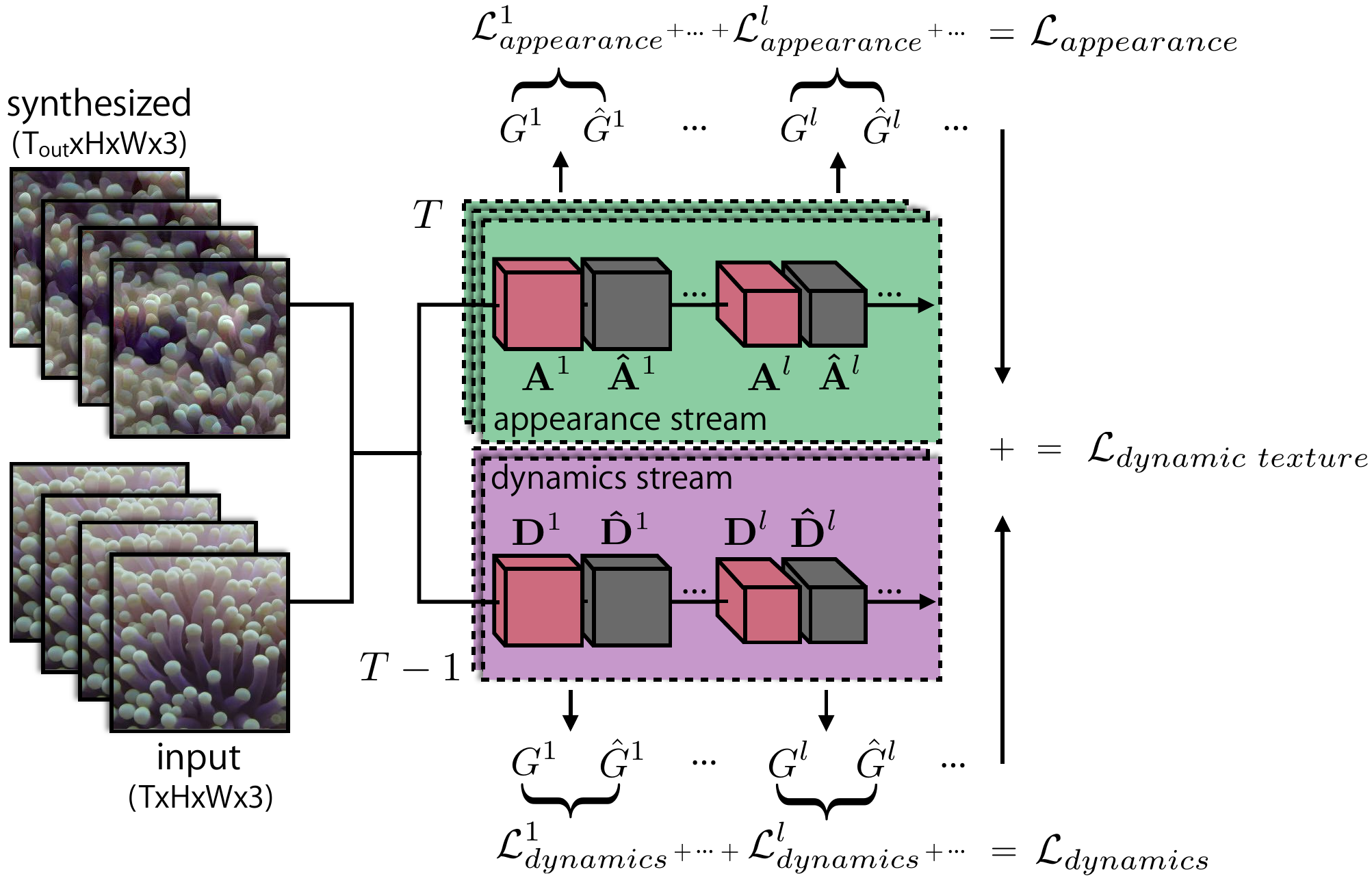, width = 0.48\textwidth}
\end{center}
\vspace{-0.45cm}
\caption{Two-stream dynamic texture generation.
Sets of Gram matrices represent a texture's appearance and 
dynamics.
Matching these statistics allows for the generation of novel
textures as well as style transfer between textures.}
\label{fig:architecture}
\end{figure}

Our proposed two-stream approach consists of an appearance
stream, representing the static (texture) appearance of each frame,
and a dynamics stream, representing temporal 
variations between frames.
Each stream consists of a ConvNet whose activation 
statistics are used to characterize the dynamic texture.
Synthesizing a dynamic texture is formulated as an optimization 
problem with the objective of matching the activation 
statistics.
Our dynamic texture synthesis approach is summarized in Fig.\ \ref{fig:architecture}
and the individual pieces are described in turn in the
following sections.

\subsection{Texture model: Appearance stream}
The appearance stream follows the spatial texture model
introduced by Gatys et al.\ \cite{gatys2015} which we
briefly review here.
The key idea is that feature correlations in a 
ConvNet trained for 
object recognition  
capture texture appearance.
We use the same publicly available normalized VGG-19 network \cite{simonyan2014very} used by Gatys et al.\ \cite{gatys2015}.

To capture the appearance of an input dynamic texture, we first
perform a forward pass with each frame of the image sequence
through the ConvNet and compute the feature activations,
$\mathbf{A}^{lt} \in \mathbb{R}^{N_l\times M_l}$, for various
levels in the network, where $N_l$ and $M_l$ denote
the number of filters and the number of spatial locations of layer
$l$ at time $t$, respectively.
The correlations of the filter responses in a particular layer are
averaged over the frames and encapsulated by a Gram matrix,
$\mathbf{G}^{l} \in \mathbb{R}^{N_l \times N_l}$, whose
entries are given by
$G_{ij}^l = \frac{1}{T N_l M_l} \sum_{t=1}^T \sum_{k=1}^{M_l} A_{ik}^{lt} A_{jk}^{lt}$,
where $T$ denotes the number of input frames
and $A_{ik}^{lt}$ denotes the activation of feature $i$ at
location $k$ in layer $l$ on the target frame $t$.
The synthesized texture appearance is similarly represented by a
Gram matrix, $\hat{\mathbf{G}}^{lt} \in \mathbb{R}^{N_l \times N_l}$,
whose activations are given by
$\hat{G}_{ij}^{lt} = \frac{1}{N_l M_l} \sum_{k=1}^{M_l} \hat{A}_{ik}^{lt} \hat{A}_{jk}^{lt}$,
where $\hat{A}_{ik}^{lt}$ denotes the activation of feature $i$ at
location $k$ in layer $l$ on the synthesized frame $t$.
The appearance loss, $\mathcal{L}_\text{appearance}$, is then 
defined as the temporal average of the mean squared error between
the Gram matrix of the input texture and that of the generated
texture computed at each frame:
\begin{equation}
   \mathcal{L}_\text{appearance} = \frac{1}{L_\text{app} T_\text{out}} \sum_{t=1}^{T_\text{out}} \sum_{l} \Vert \mathbf{G}^l - \hat{\mathbf{G}}^{lt} \Vert^2_F\ ,
   \label{eq:apploss}
\end{equation}
where $L_\text{app}$ is the number of layers used to compute Gram
matrices, $T_\text{out}$ is the number of frames being generated in
the output, and $\Vert \cdot \Vert_F$ is the Frobenius norm.
Consistent with previous work \cite{gatys2015}, we compute Gram matrices on the
following layers: 
\emph{conv1\_1}, \emph{pool1}, \emph{pool2}, \emph{pool3}, and \emph{pool4}.

\subsection{Texture model: Dynamics stream}

There are three primary goals in designing our dynamics stream.
First, the activations of the network must 
represent the temporal variation of the input pattern.
Second, the activations should be largely invariant to the
appearance of the images which should be characterized
by the appearance stream described above.
Finally, the representation must be differentiable to enable 
synthesis.
By analogy to the appearance stream, an obvious choice
is a ConvNet architecture suited for computing
optical flow (\eg, \cite{dosovitskiy2015,ilg2017}) which
is naturally differentiable.
However, with most such models it is unclear how invariant
their layers are to appearance.
Instead, we propose a novel network architecture which is
motivated by the spacetime-oriented energy model
\cite{derpanis2012spacetime,simoncelli1998}.

In motion energy models, the velocity of image content (\ie, motion)
is interpreted as a three-dimensional orientation in the $x$-$y$-$t$
spatiotemporal domain
\cite{adelson1985spatiotemporal,fahle1981,heeger1988,simoncelli1998,watson1983}.
In the frequency domain, the signal energy of a translating
pattern can be shown to lie on a plane through the origin
where the slant of the plane is defined by the velocity of
the  pattern.
Thus, motion energy models attempt to identify this 
orientation-plane (and hence the patterns velocity) via
a set of image filtering operations.
More generally
the constituent
spacetime orientations for a spectrum of common
visual patterns (including translation and dynamic
textures) can serve as a basis for describing the temporal
variation of an image sequence \cite{derpanis2012spacetime}.
This suggests that motion energy models may form an
ideal basis for our dynamics stream.

Specifically, we use the spacetime-oriented energy model
\cite{derpanis2012spacetime,simoncelli1998} to motivate our
network architecture which we briefly review here; 
see \cite{derpanis2012spacetime} for a more in-depth 
description.
Given an input video, 
a bank of oriented 3D
filters are applied which are sensitive to a range of
spatiotemporal orientations.
These filter activations are rectified (squared) and
pooled over local regions to make the responses robust
to the phase of the input signal, \ie, robust to the
alignment of the filter with the underlying image
structure.
Next, filter activations consistent with the same spacetime
orientation are summed.
These responses provide a pixelwise distributed measure
of which orientations (frequency domain planes) are
present in the input.
However, these responses are confounded by local image
contrast that makes 
it difficult to determine
whether a high response is indicative of the presence of
a spacetime orientation or simply due to high image
contrast.
To address this ambiguity, an $\textrm{L}_1$
normalization is applied across orientation responses which
results in a representation that is robust to local
appearance variations but highly selective to 
spacetime orientation.

Using this model as our basis, we propose the following 
fully convolutional 
network
\cite{shelhamer2017}.
Our ConvNet input is a pair of temporally consecutive greyscale images.
Each input pair is first normalized to have zero-mean and unit
variance.
This step provides a level of invariance to overall
brightness and contrast, \ie, global additive and
multiplicative signal variations.
The first layer consists of 32 3D spacetime convolution
filters of size $11\times 11 \times 2$
($\text{height} \times \text{width} \times \text{time}$).
Next, a squaring activation function and $5 \times 5$
spatial max-pooling (with a stride of one) is applied to
make the responses robust to local signal phase.
A $1\times 1$
convolution layer follows with 64 filters that combines
energy measurements that are consistent
with the same orientation.
Finally, to remove local contrast dependence, an
$\text{L}_1$ divisive normalization is applied.

To capture spacetime orientations beyond those capable
with the limited receptive fields used in the initial
layer, we compute a five-level spatial Gaussian pyramid.
Each pyramid level is
processed independently
with the same spacetime-oriented energy model and then
bilinearly upsampled to the original resolution and
concatenated.

\begin{figure}[t]
\begin{center}
    \epsfig{file=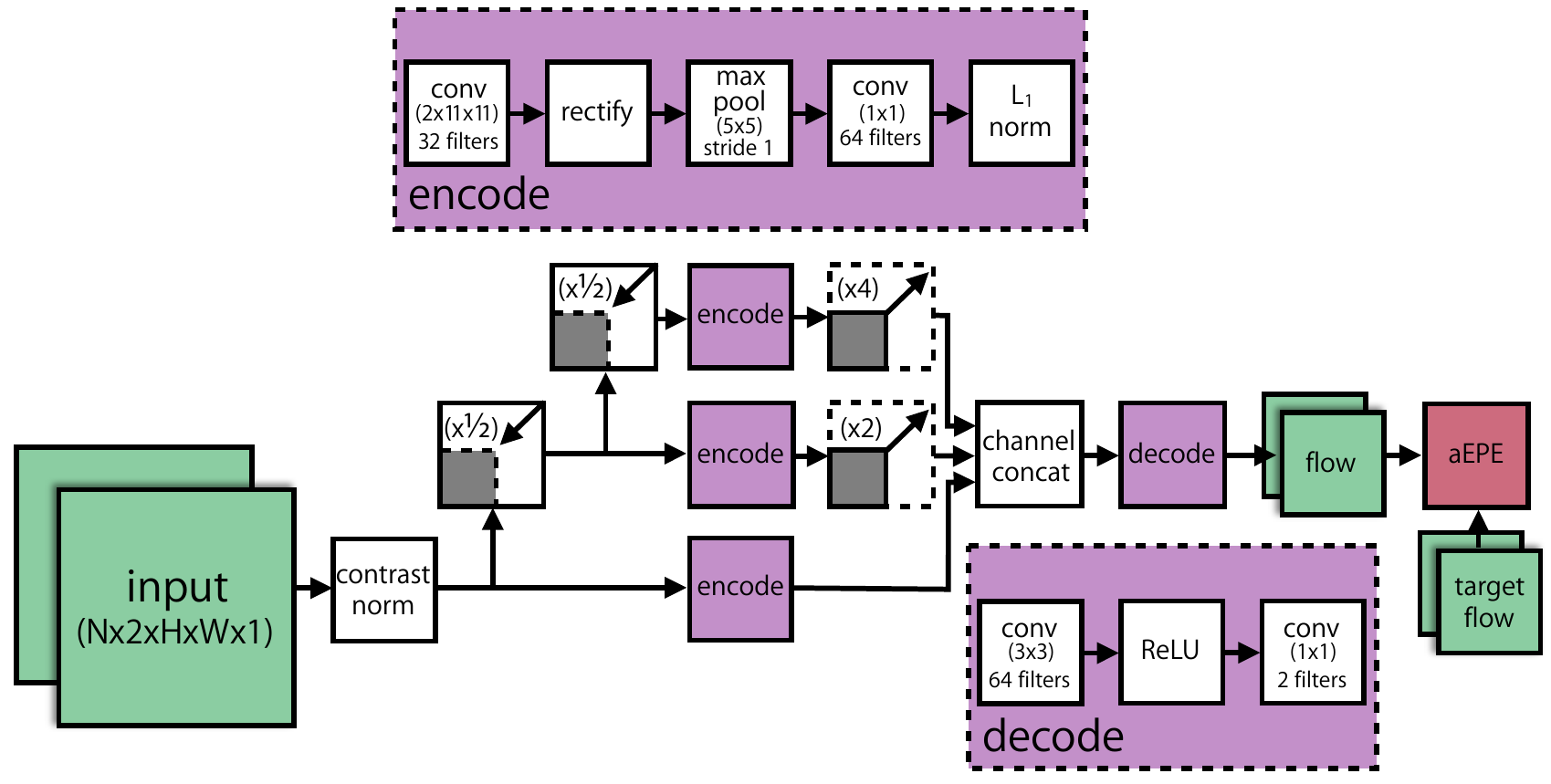, width = 0.48\textwidth}
\end{center}
\vspace{-0.45cm}
\caption{Dynamics stream ConvNet. The ConvNet is based on a
spacetime-oriented energy model
\cite{derpanis2012spacetime, simoncelli1998} and is trained
for optical flow prediction.
Three scales are shown for illustration;
in practice five scales were used.}
\label{fig:MSOE}
\end{figure}

Prior energy model instantiations (\eg,
\cite{adelson1985spatiotemporal,derpanis2012spacetime,simoncelli1998})
used handcrafted filter weights.
While a similar approach could be followed here, we
opt to learn the weights so that they are better
tuned to natural imagery.
To train the network weights, we add additional decoding
layers that take the concatenated distributed
representation and apply a $3\times 3$ convolution
(with 64 filters), ReLU activation, and a $1\times 1$
convolution (with 2 filters) that yields a two channel
output encoding the optical flow directly.
The proposed architecture is illustrated in
Fig.\ \ref{fig:MSOE}.

For training, we use the standard average
endpoint error (aEPE) flow metric (\ie, $\text{L}_2$
norm) between the predicted flow and the ground truth
flow as the loss.
Since no large-scale flow dataset exists that captures
natural imagery with groundtruth flow, we take an
unlabeled video dataset and apply an existing flow
estimator \cite{revaud2015epicflow} to estimate optical
flow for training,
\cf \cite{tran2016}.
For training data, we used videos from the UCF101
dataset \cite{soomro2012ucf101} with geometric
and photometric data augmentations similar to those used
by FlowNet \cite{dosovitskiy2015}, and optimized the aEPE loss using
Adam \cite{kingma2017}.
Inspection of the learned filters in the initial layer
showed evidence of spacetime-oriented filters, consistent with
the handcrafted filters used in previous work \cite{derpanis2012spacetime}.

Similar to the appearance stream, filter response correlations
in a particular layer of the dynamics
stream are averaged over the number of image frame
pairs and encapsulated by a Gram matrix,
$\mathbf{G}^{l} \in \mathbb{R}^{N_l \times N_l}$,
whose entries are given by
$G_{ij}^l = \frac{1}{(T-1) N_l M_l} \sum_{t=1}^{T-1} \sum_{k=1}^{M_l} D_{ik}^{lt} D_{jk}^{lt}$,
where $D_{ik}^{lt}$ denotes the activation of feature $i$ at
location $k$ in layer $l$ on the target frames $t$ and $t+1$.
The dynamics of the synthesized texture is represented
by a Gram matrix of filter response correlations 
computed separately for each pair of frames,
$\hat{\mathbf{G}}^{lt} \in \mathbb{R}^{N_l \times N_l}$,
with entries
$\hat{G}_{ij}^{lt} = \frac{1}{N_l M_l} \sum_{k=1}^{M_l} \hat{D}_{ik}^{lt} \hat{D}_{jk}^{lt}$,
where $\hat{D}_{ik}^{lt}$ denotes the activation of feature $i$ at
location $k$ in layer $l$ on the synthesized frames $t$ and $t+1$.
The dynamics loss, $\mathcal{L}_\text{dynamics}$, is defined as
the average of the mean squared error between the Gram matrices
of the input texture
and those of the generated texture:
\begin{equation}
   \mathcal{L}_\text{dynamics} = \frac{1}{L_\text{dyn} (T_\text{out}-1)}\sum_{t=1}^{T_\text{out}-1} \sum_{l}  \Vert \mathbf{G}^l - \hat{\mathbf{G}}^{lt}\Vert^2_F, \label{eq:dynloss}
\end{equation}
where $L_\text{dyn}$ is the number of ConvNet layers being used
in the dynamics stream.

Here we propose to use the output of the concatenation layer,
where the multiscale distributed representation of orientations is
stored, as the layer to compute the Gram matrix.
While it is tempting to use the predicted flow output from the
network, this generally yields poor results as shown in our evaluation.
Due to the complex, temporal variation present in dynamic
textures, they contain a variety of local spacetime
orientations rather than a single dominant orientation.
As a result, the flow estimates will tend to be an average of the
underlying  orientation measurements and consequently not
descriptive. A comparison between the texture synthesis results using the concatenation layer and the predicted flow output is provided in Sec.\ \ref{empirical_evaluation}.

\subsection{Texture generation}\label{sec:texgen}
The overall dynamic texture loss consists of the combination of the appearance loss, Eq.\ (\ref{eq:apploss}),
and the dynamics loss, Eq.\ (\ref{eq:dynloss}):
\begin{equation}
   \mathcal{L}_\text{dynamic texture} = \alpha\mathcal{L}_\text{appearance} + \beta \mathcal{L}_\text{dynamics}, \label{eq:dyntexloss}
\end{equation}
where $\alpha$ and $\beta$ are the weighting factors for the
appearance and dynamics content, respectively.
Dynamic textures are implicitly defined as the (local) minima 
of this loss.
Textures are generated by optimizing Eq.\ 
(\ref{eq:dyntexloss}) with respect to the spacetime volume,
\ie, the pixels of the video.
Variations in the resulting texture are found by initializing the
optimization process using IID Gaussian noise.
Consistent with previous work \cite{gatys2015}, we use
L-BFGS \cite{liu1989} optimization. 

Naive application of the outlined approach will consume
increasing amounts of memory as the temporal extent of the 
dynamic texture grows; this makes it impractical to generate
longer sequences.
Instead, long sequences can be incrementally generated by
separating the sequence into subsequences and optimizing them 
sequentially.  This is realized by initializing the first frame of a subsequence as the last 
frame from the previous subsequence and keeping it fixed throughout
the optimization.
The remaining frames of the subsequence are initialized randomly and
optimized as above.
This ensures temporal consistency across synthesized subsequences
and can be viewed as a form of coordinate descent for the full
sequence objective.
The flexibility of this framework allows other texture generation
problems to be handled simply by altering the 
initialization of frames and controlling which
frames or frame regions are updated.

\begin{figure*}[t]
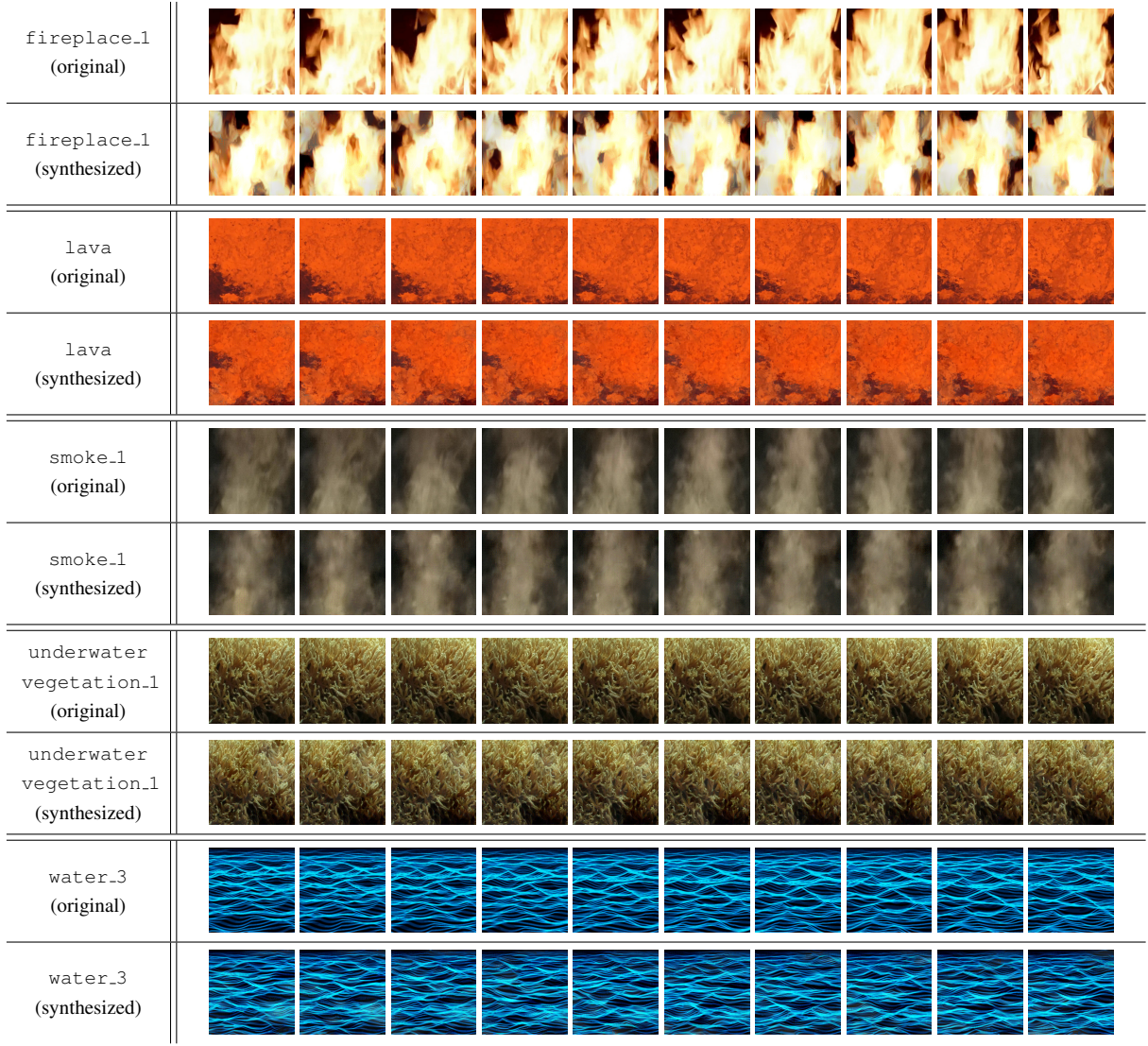

\begin{center}
\begin{tabular}{ >{\centering\arraybackslash} m{0.74in} || >{\centering\arraybackslash} m{5.2in} }

{\footnotesize \texttt{fireplace\_1}\break(original)} &
\showtexture{fireplace_1/frame_} \\
\hline
{\footnotesize \texttt{fireplace\_1}\break(synthesized)} &
\showtexture{fireplace_1_output/frame_} \\

\hline \hline
{\footnotesize \texttt{lava}\break(original)} &
\showtexture{lava/frame_} \\
\hline
{\footnotesize \texttt{lava}\break(synthesized)} &
\showtexture{lava_output/frame_} \\

\hline \hline
{\footnotesize \texttt{smoke\_1}\break(original)} &
\showtexture{smoke_1/frame_} \\
\hline
{\footnotesize \texttt{smoke\_1}\break(synthesized)} &
\showtexture{smoke_1_output/frame_} \\

\hline \hline
{\footnotesize \texttt{underwater}\break\texttt{vegetation\_1}\break(original)} &
\showtexture{underwater_vegetation/frame_} \\
\hline
{\footnotesize \texttt{underwater}\break\texttt{vegetation\_1}\break (synthesized)} &
\showtexture{underwater_vegetation_output/frame_} \\

\hline \hline
{\footnotesize \texttt{water\_3}\break(original)} & 
\showtexture{water_3/frame_} \\
\hline
{\footnotesize \texttt{water\_3}\break(synthesized)} & 
\showtexture{water_3_output/frame_} \\
\end{tabular}
\end{center}
\vspace{-0.45cm}
\caption{Dynamic texture synthesis success examples. Names correspond
to files in the supplemental material.}
 \label{fig:successes}
\end{figure*}

\section{Experimental results}
\label{empirical_evaluation}
The goal of (dynamic) texture synthesis is to generate 
samples that are indistinguishable from the real input target
texture by a human observer.
In this section, we present a variety of synthesis results
including a user study to quantitatively evaluate the realism
of our results.
Given their temporal nature, our results are best viewed as 
videos.
Our two-stream architecture was implemented using TensorFlow
\cite{tabadi2015tensorflowlong}.
Results were generated using an NVIDIA Titan X (Pascal) GPU
and synthesis times ranged between one to three hours 
to generate $12$ frames with an image resolution of 
$256 \times 256$.
For our full synthesis results and source code, please refer to the
supplemental material on the project website: \url{ryersonvisionlab.github.io/two-stream-projpage}.

\subsection{Dynamic texture synthesis}
\label{eval_dynamictexturesynthesis}
We applied our dynamic texture synthesis process 
to a wide range of textures which were selected from the 
DynTex \cite{peteri2010} database and others we collected in
the wild.
Included in our supplemental material are synthesized results
of nearly 60 different textures that encapsulate a range of
phenomena, such as flowing water, waves, clouds, fire, rippling
flags, waving plants, and schools of fish.
Some sample frames are shown in Fig.\ \ref{fig:successes}
but we encourage readers to view the videos to fully appreciate
the results.
In addition, we performed a comparison with \cite{funke2017} and 
\cite{xie2017synthesizing}.
Generally, we found our results to be qualitatively comparable or better than
these methods.
See the supplemental for more details on the comparisons with these methods.

We also generated dynamic textures incrementally, as described in
Sec.\ \ref{sec:texgen}.
The resulting textures were perceptually indistinguishable from those
generated with the batch process.
Another extension that we explored were textures with no 
discernible temporal seam between the last and first frames.
Played as a loop, these textures appear to be temporally endless.
This was achieved by assuming that the first frame follows the
final frame and adding an additional loss for the dynamics 
stream evaluated on that pair of frames.

\begin{figure}[t]
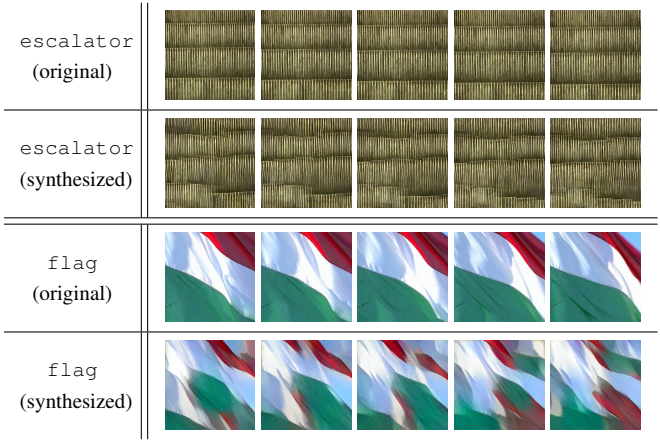

\begin{center}
\begin{tabular}{ >{\centering\arraybackslash} m{0.55in} || >{\centering\arraybackslash} m{2.50in} }
{\footnotesize \texttt{escalator}\break(original)} & 
\showtextureshort{escalator/frame_} \\
\hline
{\footnotesize \texttt{escalator}\break(synthesized)} & 
\showtextureshort{escalator_output/frame_} \\
\hline \hline
{\footnotesize \texttt{flag}\break(original)} &
\showtextureshort{flag/frame_} \\
\hline
{\footnotesize \texttt{flag}\break(synthesized)} &
\showtextureshort{flag_output/frame_} \\
\end{tabular}
\end{center}
\vspace{-0.45cm}
\caption{Dynamic texture synthesis failure examples. In
these cases, the failures are attributed to either the
appearance or the dynamics not being homogeneous.}
\label{fig:failures}
\end{figure}

Example failure modes of our method are presented in Fig.\ 
\ref{fig:failures}.
In general, we find that most failures result from inputs that
violate the underlying assumption of a dynamic texture, \ie, 
the appearance and/or dynamics are not spatiotemporally homogeneous.
In the case of the \texttt{escalator} example, the long edge 
structures in the appearance are not spatially homogeneous, 
and the dynamics vary due to perspective effects that
change the motion from downward to outward.
The resulting synthesized texture captures an overall downward 
motion but lacks the perspective effects and is unable to 
consistently reproduce the long edge structures.
This is consistent with previous observations
on static texture synthesis \cite{gatys2015} and suggests it is a 
limitation of the appearance stream.

Another example is the \texttt{flag} sequence where the rippling 
dynamics are relatively homogeneous across the pattern but the 
appearance  varies spatially.
As expected, the generated texture does not faithfully
reproduce the appearance; however, it does exhibit plausible 
rippling dynamics.
In the supplemental material, we include an additional failure 
case, \texttt{cranberries}, which consists of a swirling pattern.
Our model faithfully reproduces the appearance
but is unable to capture the spatially varying dynamics.
Interestingly, it still produces a result
which is statistically indistinguishable from real in our user 
study discussed below.

\paragraph*{Appearance vs.\ dynamics streams}

\begin{figure}[t]
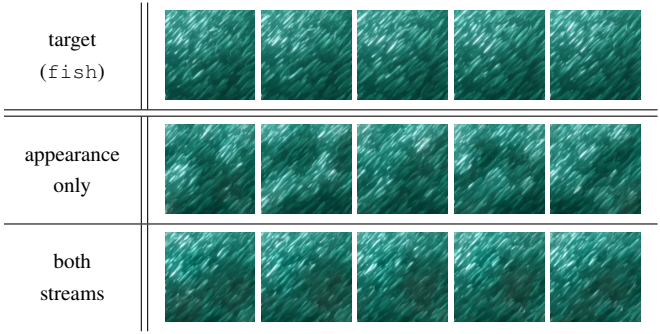

\begin{center}
\begin{tabular}{ >{\centering\arraybackslash} m{0.55in} || >{\centering\arraybackslash} m{2.50in} }
{\footnotesize target (\texttt{fish})} & 
\showtextureshort{fish/frame_} \\
\hline \hline
{\footnotesize appearance only} &
\showtextureshort{fish_spatialonly/frame_} \\
\hline
{\footnotesize both streams} & 
\showtextureshort{fish_output/frame_} \\
\end{tabular}
\end{center}
\vspace{-0.45cm}
\caption{Dynamic texture synthesis versus texture synthesis.
(top row) Target texture.
(middle)
Texture synthesis without dynamics constraints shows
consistent per-frame appearance but no temporal coherence.
(bottom)
Including both streams induces consistent appearance and dynamics.
}
\label{fig:baselines}
\end{figure}

We sought to verify that the appearance and dynamics
streams were capturing complementary information.
To validate that the texture generation of multiple frames
would not induce dynamics consistent with the input, we generated
frames starting from randomly generated noise but only using the
appearance statistics and corresponding loss, \ie,
Eq.\ \ref{eq:apploss}.
As expected, this produced frames that were valid textures but
with no coherent dynamics present.
Results for a sequence containing a school of fish are shown in
Fig.\ \ref{fig:baselines}; to examine the dynamics, see 
\texttt{fish} in the supplemental material.

Similarly, to validate that the dynamics stream did not 
inadvertently include appearance information, we generated videos
using the dynamics loss only, \ie, Eq.\ \ref{eq:dynloss}.
The resulting frames had no visible appearance and had
an extremely low dynamic range, \ie, the standard
deviation of pixel intensities was 10 for values in $[0,255]$.
This indicates a general invariance to appearance and 
suggests that our two-stream dynamic texture representation
has factored appearance and dynamics, as desired.

\subsection{User study}
Quantitative evaluation for texture synthesis is a particularly
challenging task as there is no single correct output when 
synthesizing new samples of a texture.
Like in other image generation tasks (\eg, rendering), 
human perception is ultimately the most important measure.
Thus, we performed a user study to evaluate the perceived 
realism of our synthesized textures.

Similar to previous image synthesis work (\eg, \cite{chen2017}), 
we conducted a perceptual experiment with human observers to 
quantitatively evaluate our synthesis results.
We employed a forced-choice evaluation on Amazon Mechanical
Turk (AMT) with 200 different users. Each user performed 59
pairwise comparisons between a synthesized dynamic texture and 
its target.
Users were asked to choose which appeared more realistic
after viewing the textures for an exposure time sampled
randomly from discrete intervals between 0.3 and 4.8 seconds.
Measures were taken to control the experimental conditions and
minimize the possibility of low quality data.
See the supplemental material for further experimental details
of our user study.

For comparison, we constructed a baseline by using the 
flow decode layer in the dynamics loss of Eq.\ \ref{eq:dynloss}.
This corresponds with attempting to mimic the optical flow 
statistics of the texture directly.
Textures were synthesized with this model and the user study
was repeated with an additional 200 users.
To differentiate between the models, we label ``Flow decode layer'' 
and ``Concat layer'' in the figures to describe our baseline and 
final model, respectively.

\begin{figure}[t]
	\centering
    \epsfig{file=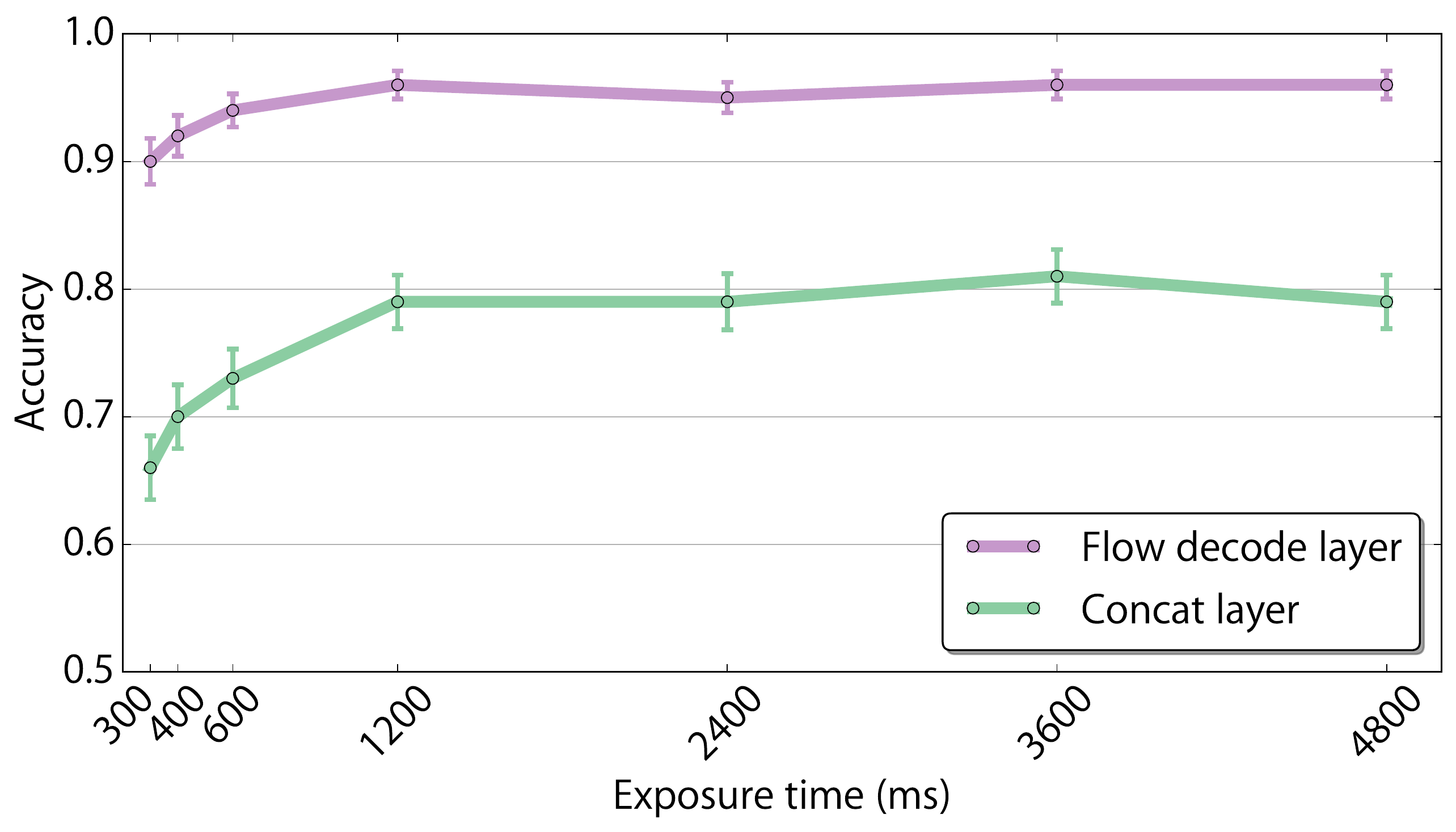, width = 0.41\textwidth}
	\caption{Time-limited pairwise comparisons across all textures with $95\%$ statistical confidence intervals.}
	\label{fig:pairwise_alltextures}
\end{figure}

The results of this study are summarized in
Fig.\ \ref{fig:pairwise_alltextures} which shows user accuracy in
differentiating real versus generated textures as a function of
time for both methods.
Overall, users are able to correctly identify the real texture
$66.1\% \pm 2.5\%$ of the time for brief 
exposures of 0.3 seconds.
This rises to $79.6\% \pm 1.1\%$ with exposures of 1.2 seconds 
and higher.
Note that ``perfect'' synthesis results would have an accuracy
of $50\%$, indicating that users were unable to differentiate 
between the real and generated textures and higher accuracy 
indicating less convincing textures.

The results clearly show that the use of the concatenation 
layer activations is far more effective than the flow decode 
layer.
This is not surprising as optical flow alone is known to be 
unreliable on many textures, particularly those with
transparency or chaotic motion (\eg, water, smoke, flames, etc.).
Also evident in these results is the time-dependant nature of 
perception for textures from both models.
Users' ability to identify the generated texture improved as 
exposure times increased to 1.2 seconds and remained relatively 
flat for longer exposures.

\begin{figure}[t]
	\centering
	\epsfig{file=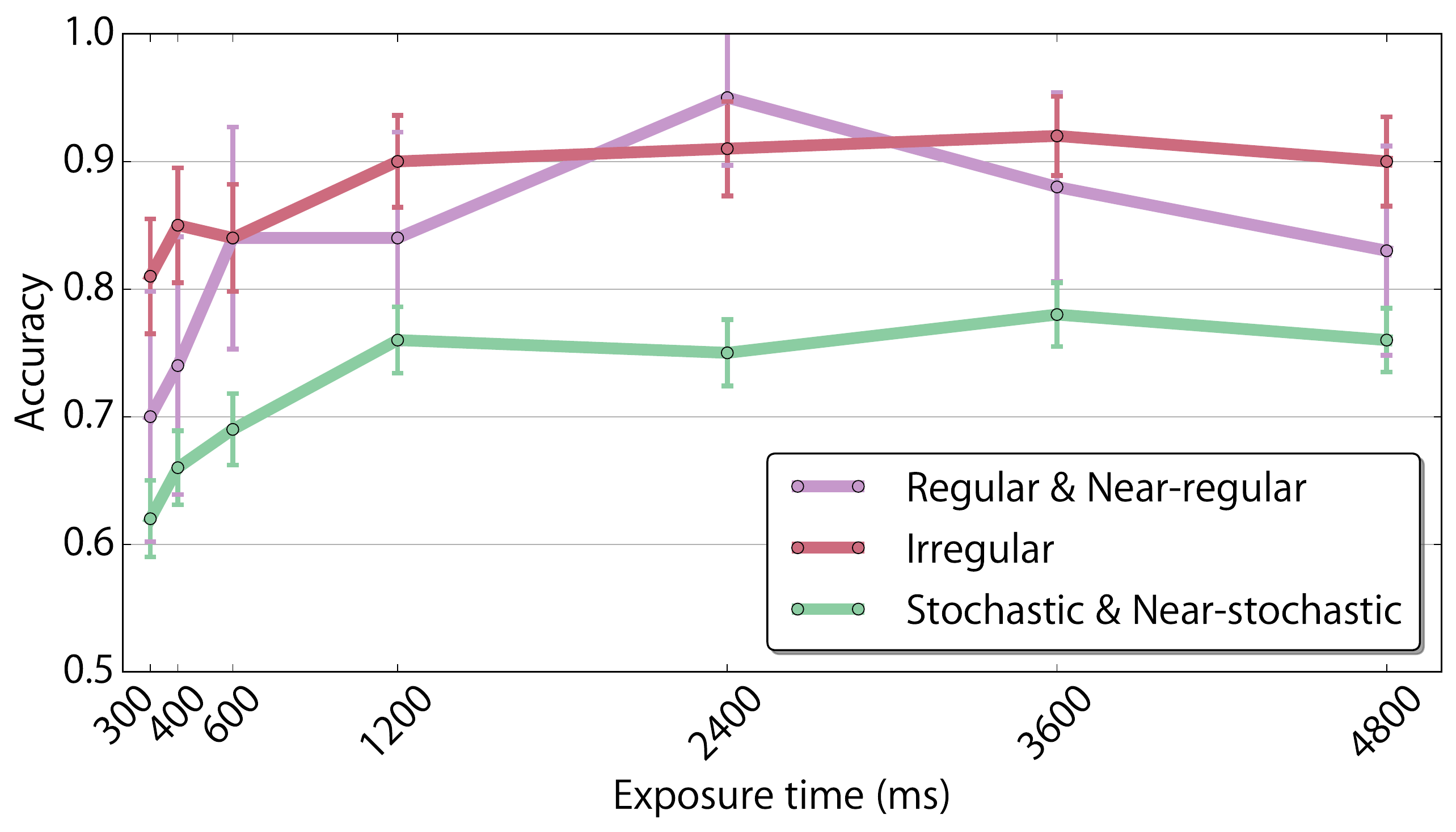, width = 0.41\textwidth}\\
    \epsfig{file=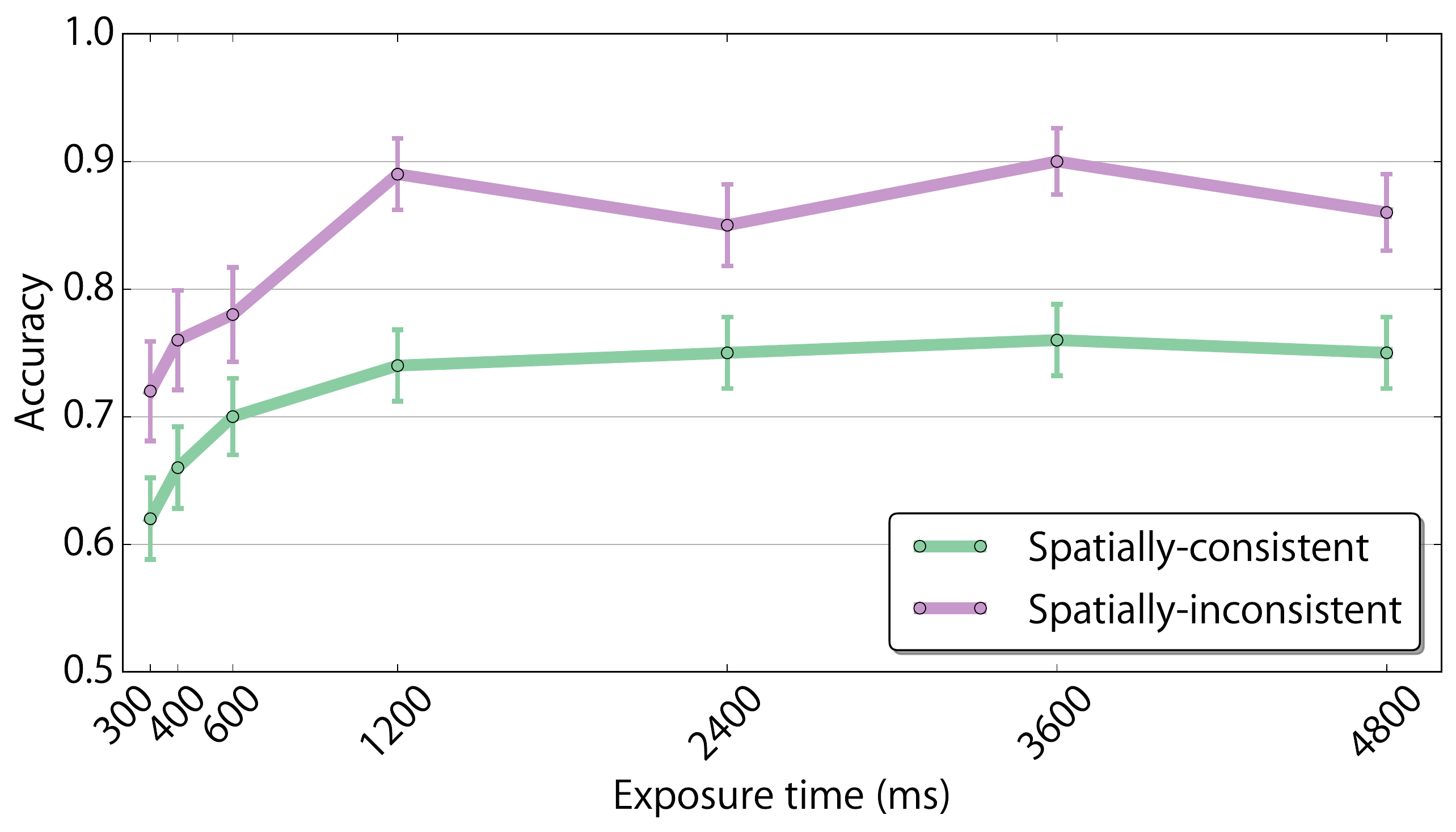, width = 0.41\textwidth}
	\caption{Time-limited pairwise comparisons across all textures, grouped by appearance (top) and dynamics (bottom).  Shown with $95\%$ statistical confidence intervals.
	}
	\label{fig:pairwise_grouped}
	\vspace{-0.2cm}
\end{figure}

To better understand the performance of our approach,
we grouped and analyzed the results in terms of
appearance and dynamics characteristics.
For appearance we used the taxonomy
presented in \cite{lin2006quantitative} and grouped textures as
either regular/near-regular (\eg, periodic tiling and brick wall), 
irregular (\eg, a field of flowers), or
stochastic/near-stochastic (\eg, tv static or water).
For dynamics we grouped textures as either 
spatially-consistent (\eg, closeup of rippling sea water) or 
spatially-inconsistent (\eg, rippling sea water juxtaposed 
with translating clouds in the sky).
Results based on these groupings can be seen in
Fig.\ \ref{fig:pairwise_grouped}.

A full breakdown of the user study results by texture and 
grouping can be found in the supplemental material.
Here we discuss some of the overall trends.
Based on appearance it is clear that textures with
large-scale  spatial consistencies (regular, near-regular, 
and irregular textures) tend to perform poorly.
Examples being \texttt{flag} and \texttt{fountain\_2} with
user accuracies of $98.9\% \pm 1.6\%$ and $90.8\% \pm 4.3\%$ 
averaged across all exposures, respectively.
This is not unexpected and is a fundamental limitation of the 
local nature of the Gram matrix representation used in the 
appearance stream which was observed in static texture synthesis 
\cite{gatys2015}.
In contrast, stochastic and near-stochastic textures 
performed significantly better as their smaller-scale local 
variations are well captured by the appearance stream, for 
instance \texttt{water\_1} and \texttt{lava} which had 
average accuracies of $53.8\% \pm 7.4\%$ and
$55.6\% \pm 7.4\%$, respectively, making them both 
statistically indistinguishable from real.

In terms of dynamics, we find that textures with
spatially-consistent dynamics (\eg, \texttt{tv\_static}, 
\texttt{water\_*}, and  \texttt{calm\_water\_*}) perform 
significantly better than those with spatially-inconsistent 
dynamics (\eg, \texttt{candle\_flame}, \texttt{fountain\_2}, 
and \texttt{snake\_*}), where the dynamics drastically differ 
across spatial locations.
For example, \texttt{tv\_static} and \texttt{calm\_water\_6}
have average accuracies of $48.6\% \pm 7.4\%$ and
$63.2\% \pm 7.2\%$, respectively, while
\texttt{candle\_flame} and \texttt{snake\_5} have average 
accuracies of $92.4\% \pm 4\%$ and $92.1\% \pm 4\%$, 
respectively.
Overall, our model is capable of reproducing a full spectrum
of spatially-consistent dynamics.
However, as the appearance shifts from containing small-scale 
spatial consistencies to containing large-scale consistencies,
performance degrades.
This was evident in the user study where the best-performing 
textures typically consisted of a stochastic or
near-stochastic appearance with spatially-consistent 
dynamics.
In contrast the worst-performing textures consisted of
regular, near-regular, or irregular appearance with
spatially-inconsistent dynamics.

\subsection{Dynamics style transfer}
The underlying assumption of our model is that appearance
and dynamics of texture can be factorized.
As such, it should allow for the transfer of the dynamics of
one texture onto the appearance of another.
This has been explored previously for artistic style transfer
\cite{champandard2016,gatys2017} with static imagery.
We accomplish this with our model by performing the same 
optimization as above, but with the target Gram matrices for 
appearance and dynamics computed from different textures.

A dynamics style transfer result is shown in Fig.\ 
\ref{fig:motiontransfer} (top), using two real videos.
Additional examples are available in the supplemental material.
We note that when performing dynamics style transfer it is important
that the appearance structure be similar in scale and semantics,
otherwise, the generated dynamic textures will look unnatural.
For instance, transferring the dynamics of a flame onto a water 
scene will generally produce implausible results.

We can also apply the dynamics of a texture to a static input image,
as the target Gram matrices for the appearance loss can be computed
on just a single frame.
This allows us to effectively animate regions of a static image.
The result of this process can be striking and is visualized in
Fig.\ \ref{fig:motiontransfer} (bottom), where the appearance is 
taken from a painting and the dynamics from a real world video.

\begin{figure}[t]
\begin{center}
\begin{tabular}{ >{\centering\arraybackslash} m{0.55in} || >{\centering\arraybackslash} m{2.50in} }
appearance target &
synthesized output \\
\hline \hline
\vspace{0.1cm}\showtexframe{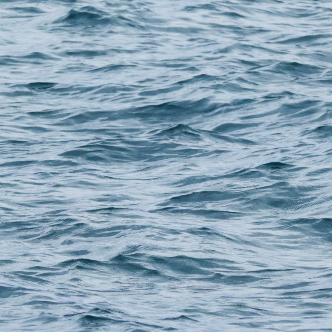} &
\showtextureshort{water_4_to_water_img_output/frame_} \\
\hline
\vspace{0.1cm}\showtexframe{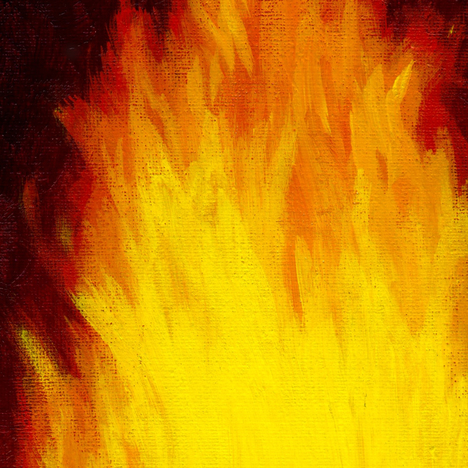} &
\showtextureshort{fireplace_1_to_fire_paint_output/frame_} \\
\end{tabular}
\end{center}
\vspace{-0.45cm}
\caption{Dynamics style transfer.
(top row) 
Appearance of still water was
used with the dynamics of a different water dynamic texture
(\texttt{water\_4}).
(bottom row) 
The appearance of a painting of fire was used
with the dynamics of a real fire (\texttt{fireplace\_1}).
Animated results and additional examples are available in
the supplemental material.} 
\label{fig:motiontransfer}
\end{figure}

\section{Discussion and summary}
In this paper, we presented a novel, two-stream model
of dynamic textures using ConvNets to represent the appearance and
dynamics.
We applied this model to a variety of dynamic texture synthesis
tasks and showed that, so long as the input textures are generally
true dynamic textures, \ie, have spatially invariant statistics and
spatiotemporally invariant dynamics, the resulting synthesized 
textures are compelling.
This was validated both qualitatively and quantitatively through
a large user study.
Further, we showed that the two-stream model enabled dynamics
style transfer, where the appearance and dynamics 
information from different sources can be combined to generate
a novel texture.

We have explored this model thoroughly and found a few limitations which we leave as directions for future work.
First, much like has been reported in recent image style transfer
work \cite{gatys2016image}, we have found that high frequency
noise and chromatic aberrations are a problem in generation.
Another issue that arises is the model fails to capture
textures with spatially-variant appearance, (\eg, 
\texttt{flag} in Fig.\ \ref{fig:failures}) and
spatially-inconsistent dynamics (\eg, \texttt{escalator} in 
Fig.\ \ref{fig:failures}).
By collapsing the local statistics into a Gram matrix, 
the spatial and temporal organization is lost.
Simple post-processing methods may alleviate some of these issues
but we believe that they also point to a need for a better
representation.
Beyond addressing these limitations, a natural next step would be
to extend the idea of a factorized representation into feed-forward
generative networks that have found success in static image
synthesis, \eg, \cite{johnson2016,ulyanov2016}.

\paragraph*{Acknowledgements}
MT is supported by a Natural Sciences and Engineering Research Council of Canada (NSERC) Canadian Graduate Scholarship. KGD and MAB are supported by NSERC Discovery Grants. This research was undertaken as part of the Vision: Science to Applications program, thanks in part to funding from the Canada First Research Excellence Fund.

\small
\bibliographystyle{ieee}
\bibliography{dynamic_texture_synthesis_main}

\appendix

\section{Experimental procedure}
Here we provide further experimental details of our user study
using Amazon Mechanical Turk (AMT). Experimental trials were
grouped into batches of Human Intelligence Tasks (HITs) for
users to complete. Each HIT consisted of 59 pairwise
comparisons between a synthesized dynamic texture and its target.
Users were asked to choose which texture appeared more realistic
after viewing each texture independently for an exposure time (in seconds) 
sampled randomly from the set \{0.3, 0.4, 0.6, 1.2, 2.4, 3.6, 4.8\}.
Note that 12 frames of the dynamic texture corresponds to 1.2 seconds, \ie, 10 frames per second.
Before viewing a dynamic texture, a centred dot is flashed twice to
indicate to the user where to look (left or right).
To prepare users for the task, the first three comparisons 
were used for warm-up, exposing them to the shortest (0.3s), 
median (1.2s), and longest (4.8s) durations.
To prevent spamming and bias, we constrained the experiment as 
follows:
users could make a choice only after both dynamic 
textures were shown;
the next texture comparison could only be made after a decision was made
for the current comparison;
a choice could not be changed after the next pair of dynamic 
textures were shown;
and users were each restricted to a single HIT.
Obvious unrealistic dynamic textures were synthesized by 
terminating synthesis early (100 iterations) and were used as sentinel tests.
Three of the 59 pairwise comparisons were sentinels and results from
users which gave incorrect answers on any of the sentinel
comparisons were not used. The left-right order of textures within a pair,
display order within a pair, and order of pairs within a HIT, were randomized.
An example of a HIT is shown in a video included with the supplemental on the project page: \texttt{HIT\_example.mp4}.

Users were paid \$2 USD per HIT, and were required to have at least
a 98\% HIT approval rating, greater than or equal to 5000 HITs
approved, and to be residing in the US. We collected results from 200 unique users
to evaluate our final model and another 200 to evaluate our baseline model.

\section{Qualitative results}
We provide videos showcasing the qualitative results of our two-stream model, including the experiments mentioned in the main manuscript, on our project page: \url{ryersonvisionlab.github.io/two-stream-projpage}. The videos are in MP4 format (H.264 codec) and are best viewed in a loop. They are enclosed in the following folders:
\begin{itemize}
	\item \texttt{target\_textures}: This folder contains the 59 dynamic textures used as targets for synthesis.
	\item \texttt{dynamic\_texture\_synthesis}: This folder contains synthesized dynamic textures where the appearance and dynamics targets are the same.
	\item \texttt{using\_concatenation\_layer}: This folder contains synthesized dynamic textures where the concatenation layer was used for computing the Gramian on the dynamics stream. These are the results from our final model.
	\item \texttt{using\_flow\_decode\_layer}: This folder contains synthesized dynamic textures where the predicted flow output is used for computing the Gramian on the dynamics stream. These are the results from our baseline.
	\item \texttt{full\_synthesis}: This folder contains regularly-synthesized dynamic textures, \ie, not incrementally-generated, nor temporally-endless, etc.
	\item \texttt{appearance\_stream\_only}: This folder contains dynamic textures synthesized using only the appearance stream of our two-stream model. The dynamics stream is not used.
	\item \texttt{incrementally\_generated}: This folder contains dynamic textures synthesized using the incremental process outlined in Section 3.3 in the main manuscript.
	\item \texttt{temporally\_endless}: This folder contains a synthesized dynamic texture (\texttt{smoke\_plume\_1}) where there is no discernible temporal seam between the last and first frames. Played as a loop, it appears to be temporally endless, thus, it is presented in animated GIF format.
	\item \texttt{dynamics\_style\_transfer}: This folder contains synthesized dynamic textures where the appearance and dynamics targets are different. Also included are videos where the synthesized dynamic texture is ``pasted'' back onto the original image it was cropped from, showing a proof-of-concept of dynamics style transfer as an artistic tool.
	\item \texttt{comparisons/funke}: This folder contains four dynamic texture synthesis comparisons between our model and a recent (unpublished) approach \cite{funke2017}. The dynamic textures chosen are those reported by Funke et al. \cite{funke2017} which exhibit spatiotemporal homogeneity. For ease of comparison, we have concatenated the results from both models with their corresponding targets.
	\item \texttt{comparisons/xie\_and\_funke}: This folder contains nine dynamic texture synthesis comparisons between our model, Funke et al.'s \cite{funke2017}, and Xie et al.'s \cite{xie2017synthesizing}. The dynamic textures chosen cover the full range of our appearance and dynamics groupings. For ease of comparison, we have concatenated the results from all models with their corresponding targets.
\end{itemize}

\section{Full user study results}
\begin{sidewaysfigure*}[t]
	\centering
	\begin{subfigure}[b]{\textwidth}
		\epsfig{file=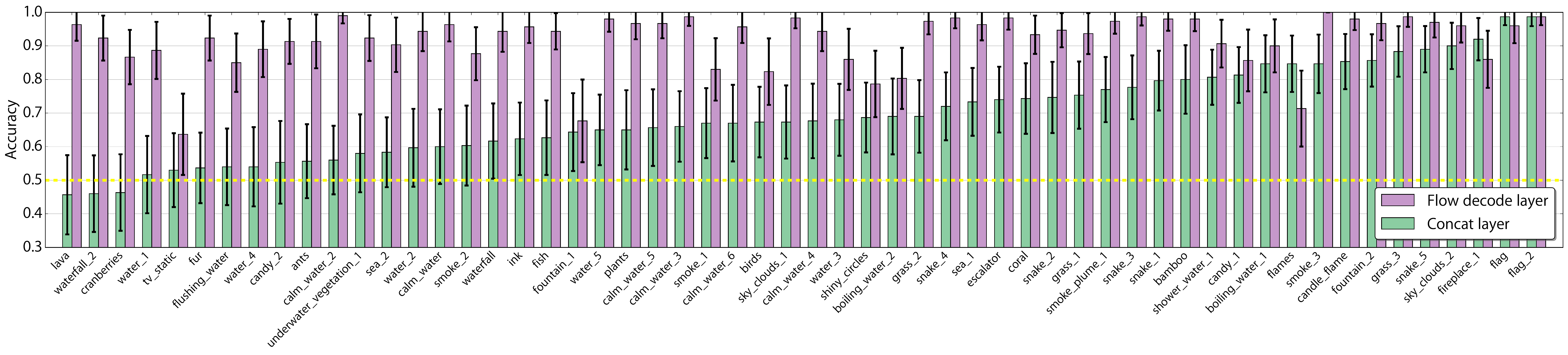, width = \textwidth}
		\vspace{-0.8cm}
		\caption{Short exposure times (300-600 ms).}
		\label{fig:alltextures_hist_short}
	\end{subfigure}
	\begin{subfigure}[b]{\textwidth}	\epsfig{file=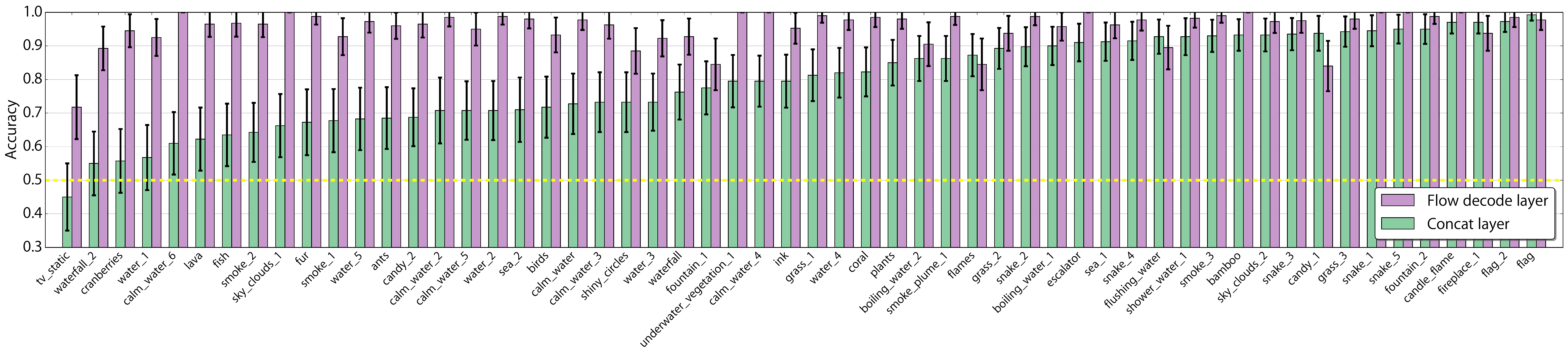, width = \textwidth}
		\vspace{-0.55cm}
		\caption{Long exposure times (1200-4800 ms).}
		\label{fig:alltextures_hist_long}
	\end{subfigure}
	\vspace{-0.65cm}
	\caption{Per-texture accuracies averaged over exposure times. Each texture accuracy includes a margin of error with a $95\%$ statistical confidence.}
	\vspace{-0.45cm}
	\label{fig:pertexture_hist}
\end{sidewaysfigure*}

Figures \ref{fig:alltextures_hist_short} and \ref{fig:alltextures_hist_long}
show histograms of the average user accuracy on each texture, averaged over a 
range of exposure times. The histogram bars are ordered from lowest
to highest accuracy, based on the results when using our final model.

Tables \ref{tab:table_concat} and \ref{tab:table_concat_range} show
the average user accuracy on each texture when using our final model.
The results are averaged over exposure times. Similarly, Tables
\ref{tab:table_flowdecode} and \ref{tab:table_flowdecode_range} show the results
when using our baseline.

Tables \ref{tab:table_concat_app} and \ref{tab:table_concat_app_range} show
the average user accuracy on texture appearance groups when using our
final model. The results are averaged over exposure times. Similarly, Tables
\ref{tab:table_flowdecode_app} and \ref{tab:table_flowdecode_app_range} show the results when using our baseline.

Tables \ref{tab:table_concat_dyn} and \ref{tab:table_concat_dyn_range} show
the average user accuracy on texture dynamics groups when using our
final model. The results are averaged over exposure times. Similarly, Tables
\ref{tab:table_flowdecode_dyn} and \ref{tab:table_flowdecode_dyn_range} show the results when using our baseline.

Tables \ref{tab:table_concat_all} and \ref{tab:table_concat_all_range} show
the average user accuracy over all textures when using our
final model. The results are averaged over exposure times. Similarly, Tables
\ref{tab:table_flowdecode_all} and \ref{tab:table_flowdecode_all_range} show the results when using our baseline.

\section{Qualitative comparisons}
We qualitatively compare our results to those of Funke et al.\ \cite{funke2017} and Xie et al.\ \cite{xie2017synthesizing}.
Note that Funke et al.\ \cite{funke2017} provided results on
only five textures and of those only four
are dynamic textures in the sense that their appearance
and dynamics are spatiotemporally coherent.
Their results on these sequences (\texttt{cranberries}, \texttt{flames}, 
\texttt{leaves}, and \texttt{water\_5}) are included in the folder
\texttt{funke} under \texttt{dynamic\_texture\_synthesis/comparisons}.
Our results are included as well.

We also compare our results to \cite{funke2017,xie2017synthesizing} on nine dynamic textures chosen to cover the full range of our dynamics and appearance groupings. We use their publicly available code and follow the parameters used in their experiments. For Funke et al.'s model \cite{funke2017}, the parameters used are $\Delta{t}=4$ and $T=12$ (recall that target dynamic textures consist of 12 frames). For the spatiotemporal and temporal models from Xie et al.\ \cite{xie2017synthesizing}, the parameters used are $T=1200$ and $\tilde{M}=3$. A comparison between our results, Funke et al.'s \cite{funke2017}, and Xie et al's \cite{xie2017synthesizing} on the nine dynamic textures are included in the folder \texttt{xie\_and\_funke} under \texttt{dynamic\_texture\_synthesis/comparisons}. Note for Xie et al.\ \cite{xie2017synthesizing}, we compare with their spatiotemporal model (labeled ``Xie et al.\ (ST)'') designed for dynamic textures with both spatial and temporal homogeneity, and their temporal model (labeled ``Xie et al.\ (FC)'') designed for dynamic textures with only temporal homogeneity.

Overall, we demonstrate that our results appear qualitatively better,
showing more temporal coherence and similarity
in dynamics and fewer artifacts, \eg, blur and flicker.
This may be a natural consequence of their limited
representation of dynamics. Although the spatiotemporal model of Xie et al.\ \cite{xie2017synthesizing} is able to synthesize dynamic textures that lack spatial homogeneity (\eg, \texttt{bamboo} and \texttt{escalator}), we note that their method can not synthesize novel dynamic textures, \ie, it appears to faithfully reproduce the target texture, reducing the applicability of their approach.

As a consequence of jointly modelling appearance and dynamics, the methods of \cite{funke2017,xie2017synthesizing} are not capable of the novel form of style transfer we demonstrated. This was enabled by the factored representation of dynamics and appearance. Furthermore, the spatiotemporal extent of the output sequence generated by Xie et al.'s \cite{xie2017synthesizing} method is limited to being equal to the input. The proposed approach does not share this limitation.

\clearpage
\onecolumn
\begin{table*}
\centering
	\resizebox{0.9\textwidth}{!}{%
	\csvreader[tabular=|l|l|,
	    table head=\hline Dynamic texture & 300 ms.\\\hline\hline,
	    late after line=\\\hline]%
	{tables/all_textures_concat_300.csv}{Name=\Name,Accuracy=\Accuracy}%
	{\Name & \Accuracy}%
	\csvreader[tabular=|l|,
	    table head=\hline 400 ms.\\\hline\hline,
	    late after line=\\\hline]%
	{tables/all_textures_concat_400.csv}{Accuracy=\Accuracy}%
	{\Accuracy}%
	\csvreader[tabular=|l|,
	    table head=\hline 600 ms.\\\hline\hline,
	    late after line=\\\hline]%
	{tables/all_textures_concat_600.csv}{Accuracy=\Accuracy}%
	{\Accuracy}%
	\csvreader[tabular=|l|,
	    table head=\hline 1200 ms.\\\hline\hline,
	    late after line=\\\hline]%
	{tables/all_textures_concat_1200.csv}{Accuracy=\Accuracy}%
	{\Accuracy}%
	\csvreader[tabular=|l|,
	    table head=\hline 2400 ms.\\\hline\hline,
	    late after line=\\\hline]%
	{tables/all_textures_concat_2400.csv}{Accuracy=\Accuracy}%
	{\Accuracy}%
	\csvreader[tabular=|l|,
	    table head=\hline 3600 ms.\\\hline\hline,
	    late after line=\\\hline]%
	{tables/all_textures_concat_3600.csv}{Accuracy=\Accuracy}%
	{\Accuracy}%
	\csvreader[tabular=|l|,
	    table head=\hline 4800 ms.\\\hline\hline,
	    late after line=\\\hline]%
	{tables/all_textures_concat_4800.csv}{Accuracy=\Accuracy}%
	{\Accuracy}%
	}
	\caption{Per-texture accuracies averaged over exposure times, using the concatenation layer. Each texture accuracy includes a margin of error with a $95\%$ statistical confidence.}
	\label{tab:table_concat}
\end{table*}

\begin{table*}
	\centering
	\resizebox{0.65\textwidth}{!}{%
	\csvreader[tabular=|l|l|,
	    table head=\hline Dynamic texture & Short (300-600 ms.)\\\hline\hline,
	    late after line=\\\hline]%
	{tables/all_textures_concat_short.csv}{Name=\Name,Accuracy=\Accuracy}%
	{\Name & \Accuracy}%
	\csvreader[tabular=|l|,
	    table head=\hline Long (1200-4800 ms.)\\\hline\hline,
	    late after line=\\\hline]%
	{tables/all_textures_concat_long.csv}{Accuracy=\Accuracy}%
	{\Accuracy}%
	\csvreader[tabular=|l|,
	    table head=\hline All (300-4800 ms.)\\\hline\hline,
	    late after line=\\\hline]%
	{tables/all_textures_concat_all.csv}{Accuracy=\Accuracy}%
	{\Accuracy}%
	}
	\caption{Per-texture accuracies averaged over a range of exposure times, using the concatenation layer. Each texture accuracy includes a margin of error with a $95\%$ statistical confidence.}
	\label{tab:table_concat_range}
\end{table*}


\begin{table*}
	\centering
	\resizebox{0.9\textwidth}{!}{%
	\csvreader[tabular=|l|l|,
	    table head=\hline Dynamic texture & 300 ms.\\\hline\hline,
	    late after line=\\\hline]%
	{tables/all_textures_flowdecode_300.csv}{Name=\Name,Accuracy=\Accuracy}%
	{\Name & \Accuracy}%
	\csvreader[tabular=|l|,
	    table head=\hline 400 ms.\\\hline\hline,
	    late after line=\\\hline]%
	{tables/all_textures_flowdecode_400.csv}{Accuracy=\Accuracy}%
	{\Accuracy}%
	\csvreader[tabular=|l|,
	    table head=\hline 600 ms.\\\hline\hline,
	    late after line=\\\hline]%
	{tables/all_textures_flowdecode_600.csv}{Accuracy=\Accuracy}%
	{\Accuracy}%
	\csvreader[tabular=|l|,
	    table head=\hline 1200 ms.\\\hline\hline,
	    late after line=\\\hline]%
	{tables/all_textures_flowdecode_1200.csv}{Accuracy=\Accuracy}%
	{\Accuracy}%
	\csvreader[tabular=|l|,
	    table head=\hline 2400 ms.\\\hline\hline,
	    late after line=\\\hline]%
	{tables/all_textures_flowdecode_2400.csv}{Accuracy=\Accuracy}%
	{\Accuracy}%
	\csvreader[tabular=|l|,
	    table head=\hline 3600 ms.\\\hline\hline,
	    late after line=\\\hline]%
	{tables/all_textures_flowdecode_3600.csv}{Accuracy=\Accuracy}%
	{\Accuracy}%
	\csvreader[tabular=|l|,
	    table head=\hline 4800 ms.\\\hline\hline,
	    late after line=\\\hline]%
	{tables/all_textures_flowdecode_4800.csv}{Accuracy=\Accuracy}%
	{\Accuracy}%
	}
	\caption{Per-texture accuracies averaged over exposure times, using the flow decode layer. Each texture accuracy includes a margin of error with a $95\%$ statistical confidence.}
	\label{tab:table_flowdecode}
\end{table*}

\begin{table*}
	\centering
	\resizebox{0.65\textwidth}{!}{%
	\csvreader[tabular=|l|l|,
	    table head=\hline Dynamic texture & Short (300-600 ms.)\\\hline\hline,
	    late after line=\\\hline]%
	{tables/all_textures_flowdecode_short.csv}{Name=\Name,Accuracy=\Accuracy}%
	{\Name & \Accuracy}%
	\csvreader[tabular=|l|,
	    table head=\hline Long (1200-4800 ms.)\\\hline\hline,
	    late after line=\\\hline]%
	{tables/all_textures_flowdecode_long.csv}{Accuracy=\Accuracy}%
	{\Accuracy}%
	\csvreader[tabular=|l|,
	    table head=\hline All (300-4800 ms.)\\\hline\hline,
	    late after line=\\\hline]%
	{tables/all_textures_flowdecode_all.csv}{Accuracy=\Accuracy}%
	{\Accuracy}%
	}
	\caption{Per-texture accuracies averaged over a range of exposure times, using the flow decode layer. Each texture accuracy includes a margin of error with a $95\%$ statistical confidence.}
	\label{tab:table_flowdecode_range}
\end{table*}


\clearpage
\begin{table*}
	\centering
	\resizebox{0.9\textwidth}{!}{%
	\csvreader[tabular=|l|l|,
	    table head=\hline Appearance group & 300 ms.\\\hline\hline,
	    late after line=\\\hline]%
	{tables/appearance_grouping_concat_300.csv}{Name=\Name,Accuracy=\Accuracy}%
	{\Name & \Accuracy}%
	\csvreader[tabular=|l|,
	    table head=\hline 400 ms.\\\hline\hline,
	    late after line=\\\hline]%
	{tables/appearance_grouping_concat_400.csv}{Accuracy=\Accuracy}%
	{\Accuracy}%
	\csvreader[tabular=|l|,
	    table head=\hline 600 ms.\\\hline\hline,
	    late after line=\\\hline]%
	{tables/appearance_grouping_concat_600.csv}{Accuracy=\Accuracy}%
	{\Accuracy}%
	\csvreader[tabular=|l|,
	    table head=\hline 1200 ms.\\\hline\hline,
	    late after line=\\\hline]%
	{tables/appearance_grouping_concat_1200.csv}{Accuracy=\Accuracy}%
	{\Accuracy}%
	\csvreader[tabular=|l|,
	    table head=\hline 2400 ms.\\\hline\hline,
	    late after line=\\\hline]%
	{tables/appearance_grouping_concat_2400.csv}{Accuracy=\Accuracy}%
	{\Accuracy}%
	\csvreader[tabular=|l|,
	    table head=\hline 3600 ms.\\\hline\hline,
	    late after line=\\\hline]%
	{tables/appearance_grouping_concat_3600.csv}{Accuracy=\Accuracy}%
	{\Accuracy}%
	\csvreader[tabular=|l|,
	    table head=\hline 4800 ms.\\\hline\hline,
	    late after line=\\\hline]%
	{tables/appearance_grouping_concat_4800.csv}{Accuracy=\Accuracy}%
	{\Accuracy}%
	}
	\caption{Accuracies of textures grouped by appearances, averaged over exposure times, using the concatenation layer. Each texture accuracy includes a margin of error with a $95\%$ statistical confidence.}
	\label{tab:table_concat_app}
\end{table*}

\begin{table*}
	\centering
	\resizebox{0.65\textwidth}{!}{%
	\csvreader[tabular=|l|l|,
	    table head=\hline Appearance group & Short (300-600 ms.)\\\hline\hline,
	    late after line=\\\hline]%
	{tables/appearance_grouping_concat_short.csv}{Name=\Name,Accuracy=\Accuracy}%
	{\Name & \Accuracy}%
	\csvreader[tabular=|l|,
	    table head=\hline Long (1200-4800 ms.)\\\hline\hline,
	    late after line=\\\hline]%
	{tables/appearance_grouping_concat_long.csv}{Accuracy=\Accuracy}%
	{\Accuracy}%
	\csvreader[tabular=|l|,
	    table head=\hline All (300-4800 ms.)\\\hline\hline,
	    late after line=\\\hline]%
	{tables/appearance_grouping_concat_all.csv}{Accuracy=\Accuracy}%
	{\Accuracy}%
	}
	\caption{Accuracies of textures grouped by appearances, averaged over a range of exposure times, using the concatenation layer. Each texture accuracy includes a margin of error with a $95\%$ statistical confidence.}
	\label{tab:table_concat_app_range}
\end{table*}

\begin{table*}
	\centering
	\resizebox{0.9\textwidth}{!}{%
	\csvreader[tabular=|l|l|,
	    table head=\hline Appearance group & 300 ms.\\\hline\hline,
	    late after line=\\\hline]%
	{tables/appearance_grouping_flowdecode_300.csv}{Name=\Name,Accuracy=\Accuracy}%
	{\Name & \Accuracy}%
	\csvreader[tabular=|l|,
	    table head=\hline 400 ms.\\\hline\hline,
	    late after line=\\\hline]%
	{tables/appearance_grouping_flowdecode_400.csv}{Accuracy=\Accuracy}%
	{\Accuracy}%
	\csvreader[tabular=|l|,
	    table head=\hline 600 ms.\\\hline\hline,
	    late after line=\\\hline]%
	{tables/appearance_grouping_flowdecode_600.csv}{Accuracy=\Accuracy}%
	{\Accuracy}%
	\csvreader[tabular=|l|,
	    table head=\hline 1200 ms.\\\hline\hline,
	    late after line=\\\hline]%
	{tables/appearance_grouping_flowdecode_1200.csv}{Accuracy=\Accuracy}%
	{\Accuracy}%
	\csvreader[tabular=|l|,
	    table head=\hline 2400 ms.\\\hline\hline,
	    late after line=\\\hline]%
	{tables/appearance_grouping_flowdecode_2400.csv}{Accuracy=\Accuracy}%
	{\Accuracy}%
	\csvreader[tabular=|l|,
	    table head=\hline 3600 ms.\\\hline\hline,
	    late after line=\\\hline]%
	{tables/appearance_grouping_flowdecode_3600.csv}{Accuracy=\Accuracy}%
	{\Accuracy}%
	\csvreader[tabular=|l|,
	    table head=\hline 4800 ms.\\\hline\hline,
	    late after line=\\\hline]%
	{tables/appearance_grouping_flowdecode_4800.csv}{Accuracy=\Accuracy}%
	{\Accuracy}%
	}
	\caption{Accuracies of textures grouped by appearances, averaged over exposure times, using the flow decode layer. Each texture accuracy includes a margin of error with a $95\%$ statistical confidence.}
	\label{tab:table_flowdecode_app}
\end{table*}

\begin{table*}
	\centering
	\resizebox{0.65\textwidth}{!}{%
	\csvreader[tabular=|l|l|,
	    table head=\hline Appearance group & Short (300-600 ms.)\\\hline\hline,
	    late after line=\\\hline]%
	{tables/appearance_grouping_flowdecode_short.csv}{Name=\Name,Accuracy=\Accuracy}%
	{\Name & \Accuracy}%
	\csvreader[tabular=|l|,
	    table head=\hline Long (1200-4800 ms.)\\\hline\hline,
	    late after line=\\\hline]%
	{tables/appearance_grouping_flowdecode_long.csv}{Accuracy=\Accuracy}%
	{\Accuracy}%
	\csvreader[tabular=|l|,
	    table head=\hline All (300-4800 ms.)\\\hline\hline,
	    late after line=\\\hline]%
	{tables/appearance_grouping_flowdecode_all.csv}{Accuracy=\Accuracy}%
	{\Accuracy}%
	}
	\caption{Accuracies of textures grouped by appearances, averaged over a range of exposure times, using the flow decode layer. Each texture accuracy includes a margin of error with a $95\%$ statistical confidence.}
	\label{tab:table_flowdecode_app_range}
\end{table*}
\clearpage


\clearpage
\begin{table*}
	\centering
	\resizebox{0.9\textwidth}{!}{%
	\csvreader[tabular=|l|l|,
	    table head=\hline Dynamics group & 300 ms.\\\hline\hline,
	    late after line=\\\hline]%
	{tables/dynamics_grouping_concat_300.csv}{Name=\Name,Accuracy=\Accuracy}%
	{\Name & \Accuracy}%
	\csvreader[tabular=|l|,
	    table head=\hline 400 ms.\\\hline\hline,
	    late after line=\\\hline]%
	{tables/dynamics_grouping_concat_400.csv}{Accuracy=\Accuracy}%
	{\Accuracy}%
	\csvreader[tabular=|l|,
	    table head=\hline 600 ms.\\\hline\hline,
	    late after line=\\\hline]%
	{tables/dynamics_grouping_concat_600.csv}{Accuracy=\Accuracy}%
	{\Accuracy}%
	\csvreader[tabular=|l|,
	    table head=\hline 1200 ms.\\\hline\hline,
	    late after line=\\\hline]%
	{tables/dynamics_grouping_concat_1200.csv}{Accuracy=\Accuracy}%
	{\Accuracy}%
	\csvreader[tabular=|l|,
	    table head=\hline 2400 ms.\\\hline\hline,
	    late after line=\\\hline]%
	{tables/dynamics_grouping_concat_2400.csv}{Accuracy=\Accuracy}%
	{\Accuracy}%
	\csvreader[tabular=|l|,
	    table head=\hline 3600 ms.\\\hline\hline,
	    late after line=\\\hline]%
	{tables/dynamics_grouping_concat_3600.csv}{Accuracy=\Accuracy}%
	{\Accuracy}%
	\csvreader[tabular=|l|,
	    table head=\hline 4800 ms.\\\hline\hline,
	    late after line=\\\hline]%
	{tables/dynamics_grouping_concat_4800.csv}{Accuracy=\Accuracy}%
	{\Accuracy}%
	}
	\caption{Accuracies of textures grouped by dynamics, averaged over exposure times, using the concatenation layer. Each texture accuracy includes a margin of error with a $95\%$ statistical confidence.}
	\label{tab:table_concat_dyn}
\end{table*}

\begin{table*}
	\centering
	\resizebox{0.65\textwidth}{!}{%
	\csvreader[tabular=|l|l|,
	    table head=\hline Dynamics group & Short (300-600 ms.)\\\hline\hline,
	    late after line=\\\hline]%
	{tables/dynamics_grouping_concat_short.csv}{Name=\Name,Accuracy=\Accuracy}%
	{\Name & \Accuracy}%
	\csvreader[tabular=|l|,
	    table head=\hline Long (1200-4800 ms.)\\\hline\hline,
	    late after line=\\\hline]%
	{tables/dynamics_grouping_concat_long.csv}{Accuracy=\Accuracy}%
	{\Accuracy}%
	\csvreader[tabular=|l|,
	    table head=\hline All (300-4800 ms.)\\\hline\hline,
	    late after line=\\\hline]%
	{tables/dynamics_grouping_concat_all.csv}{Accuracy=\Accuracy}%
	{\Accuracy}%
	}
	\caption{Accuracies of textures grouped by dynamics, averaged over a range of exposure times, using the concatenation layer. Each texture accuracy includes a margin of error with a $95\%$ statistical confidence.}
	\label{tab:table_concat_dyn_range}
\end{table*}

\begin{table*}
	\centering
	\resizebox{0.9\textwidth}{!}{%
	\csvreader[tabular=|l|l|,
	    table head=\hline Dynamics group & 300 ms.\\\hline\hline,
	    late after line=\\\hline]%
	{tables/dynamics_grouping_flowdecode_300.csv}{Name=\Name,Accuracy=\Accuracy}%
	{\Name & \Accuracy}%
	\csvreader[tabular=|l|,
	    table head=\hline 400 ms.\\\hline\hline,
	    late after line=\\\hline]%
	{tables/dynamics_grouping_flowdecode_400.csv}{Accuracy=\Accuracy}%
	{\Accuracy}%
	\csvreader[tabular=|l|,
	    table head=\hline 600 ms.\\\hline\hline,
	    late after line=\\\hline]%
	{tables/dynamics_grouping_flowdecode_600.csv}{Accuracy=\Accuracy}%
	{\Accuracy}%
	\csvreader[tabular=|l|,
	    table head=\hline 1200 ms.\\\hline\hline,
	    late after line=\\\hline]%
	{tables/dynamics_grouping_flowdecode_1200.csv}{Accuracy=\Accuracy}%
	{\Accuracy}%
	\csvreader[tabular=|l|,
	    table head=\hline 2400 ms.\\\hline\hline,
	    late after line=\\\hline]%
	{tables/dynamics_grouping_flowdecode_2400.csv}{Accuracy=\Accuracy}%
	{\Accuracy}%
	\csvreader[tabular=|l|,
	    table head=\hline 3600 ms.\\\hline\hline,
	    late after line=\\\hline]%
	{tables/dynamics_grouping_flowdecode_3600.csv}{Accuracy=\Accuracy}%
	{\Accuracy}%
	\csvreader[tabular=|l|,
	    table head=\hline 4800 ms.\\\hline\hline,
	    late after line=\\\hline]%
	{tables/dynamics_grouping_flowdecode_4800.csv}{Accuracy=\Accuracy}%
	{\Accuracy}%
	}
	\caption{Accuracies of textures grouped by dynamics, averaged over exposure times, using the flow decode layer. Each texture accuracy includes a margin of error with a $95\%$ statistical confidence.}
	\label{tab:table_flowdecode_dyn}
\end{table*}

\begin{table*}
	\centering
	\resizebox{0.65\textwidth}{!}{%
	\csvreader[tabular=|l|l|,
	    table head=\hline Dynamics group & Short (300-600 ms.)\\\hline\hline,
	    late after line=\\\hline]%
	{tables/dynamics_grouping_flowdecode_short.csv}{Name=\Name,Accuracy=\Accuracy}%
	{\Name & \Accuracy}%
	\csvreader[tabular=|l|,
	    table head=\hline Long (1200-4800 ms.)\\\hline\hline,
	    late after line=\\\hline]%
	{tables/dynamics_grouping_flowdecode_long.csv}{Accuracy=\Accuracy}%
	{\Accuracy}%
	\csvreader[tabular=|l|,
	    table head=\hline All (300-4800 ms.)\\\hline\hline,
	    late after line=\\\hline]%
	{tables/dynamics_grouping_flowdecode_all.csv}{Accuracy=\Accuracy}%
	{\Accuracy}%
	}
	\caption{Accuracies of textures grouped by dynamics, averaged over a range of exposure times, using the flow decode layer. Each texture accuracy includes a margin of error with a $95\%$ statistical confidence.}
	\label{tab:table_flowdecode_dyn_range}
\end{table*}
\clearpage


\clearpage
\begin{table*}
	\centering
	\resizebox{0.9\textwidth}{!}{%
	\csvreader[tabular=|l|l|,
	    table head=\hline Group & 300 ms.\\\hline\hline,
	    late after line=\\\hline]%
	{tables/all_textures_avg_concat_300.csv}{Name=\Name,Accuracy=\Accuracy}%
	{\Name & \Accuracy}%
	\csvreader[tabular=|l|,
	    table head=\hline 400 ms.\\\hline\hline,
	    late after line=\\\hline]%
	{tables/all_textures_avg_concat_400.csv}{Accuracy=\Accuracy}%
	{\Accuracy}%
	\csvreader[tabular=|l|,
	    table head=\hline 600 ms.\\\hline\hline,
	    late after line=\\\hline]%
	{tables/all_textures_avg_concat_600.csv}{Accuracy=\Accuracy}%
	{\Accuracy}%
	\csvreader[tabular=|l|,
	    table head=\hline 1200 ms.\\\hline\hline,
	    late after line=\\\hline]%
	{tables/all_textures_avg_concat_1200.csv}{Accuracy=\Accuracy}%
	{\Accuracy}%
	\csvreader[tabular=|l|,
	    table head=\hline 2400 ms.\\\hline\hline,
	    late after line=\\\hline]%
	{tables/all_textures_avg_concat_2400.csv}{Accuracy=\Accuracy}%
	{\Accuracy}%
	\csvreader[tabular=|l|,
	    table head=\hline 3600 ms.\\\hline\hline,
	    late after line=\\\hline]%
	{tables/all_textures_avg_concat_3600.csv}{Accuracy=\Accuracy}%
	{\Accuracy}%
	\csvreader[tabular=|l|,
	    table head=\hline 4800 ms.\\\hline\hline,
	    late after line=\\\hline]%
	{tables/all_textures_avg_concat_4800.csv}{Accuracy=\Accuracy}%
	{\Accuracy}%
	}
	\caption{Average accuracy over all textures, averaged over exposure times, using the concatenation layer. Each texture accuracy includes a margin of error with a $95\%$ statistical confidence.}
	\label{tab:table_concat_all}
\end{table*}

\begin{table*}
	\centering
	\resizebox{0.65\textwidth}{!}{%
	\csvreader[tabular=|l|l|,
	    table head=\hline Group & Short (300-600 ms.)\\\hline\hline,
	    late after line=\\\hline]%
	{tables/all_textures_avg_concat_short.csv}{Name=\Name,Accuracy=\Accuracy}%
	{\Name & \Accuracy}%
	\csvreader[tabular=|l|,
	    table head=\hline Long (1200-4800 ms.)\\\hline\hline,
	    late after line=\\\hline]%
	{tables/all_textures_avg_concat_long.csv}{Accuracy=\Accuracy}%
	{\Accuracy}%
	\csvreader[tabular=|l|,
	    table head=\hline All (300-4800 ms.)\\\hline\hline,
	    late after line=\\\hline]%
	{tables/all_textures_avg_concat_all.csv}{Accuracy=\Accuracy}%
	{\Accuracy}%
	}
	\caption{Average accuracy over all textures, averaged over a range of exposure times, using the concatenation layer. Each texture accuracy includes a margin of error with a $95\%$ statistical confidence.}
	\label{tab:table_concat_all_range}
\end{table*}

\begin{table*}
	\centering
	\resizebox{0.9\textwidth}{!}{%
	\csvreader[tabular=|l|l|,
	    table head=\hline Group & 300 ms.\\\hline\hline,
	    late after line=\\\hline]%
	{tables/all_textures_avg_flowdecode_300.csv}{Name=\Name,Accuracy=\Accuracy}%
	{\Name & \Accuracy}%
	\csvreader[tabular=|l|,
	    table head=\hline 400 ms.\\\hline\hline,
	    late after line=\\\hline]%
	{tables/all_textures_avg_flowdecode_400.csv}{Accuracy=\Accuracy}%
	{\Accuracy}%
	\csvreader[tabular=|l|,
	    table head=\hline 600 ms.\\\hline\hline,
	    late after line=\\\hline]%
	{tables/all_textures_avg_flowdecode_600.csv}{Accuracy=\Accuracy}%
	{\Accuracy}%
	\csvreader[tabular=|l|,
	    table head=\hline 1200 ms.\\\hline\hline,
	    late after line=\\\hline]%
	{tables/all_textures_avg_flowdecode_1200.csv}{Accuracy=\Accuracy}%
	{\Accuracy}%
	\csvreader[tabular=|l|,
	    table head=\hline 2400 ms.\\\hline\hline,
	    late after line=\\\hline]%
	{tables/all_textures_avg_flowdecode_2400.csv}{Accuracy=\Accuracy}%
	{\Accuracy}%
	\csvreader[tabular=|l|,
	    table head=\hline 3600 ms.\\\hline\hline,
	    late after line=\\\hline]%
	{tables/all_textures_avg_flowdecode_3600.csv}{Accuracy=\Accuracy}%
	{\Accuracy}%
	\csvreader[tabular=|l|,
	    table head=\hline 4800 ms.\\\hline\hline,
	    late after line=\\\hline]%
	{tables/all_textures_avg_flowdecode_4800.csv}{Accuracy=\Accuracy}%
	{\Accuracy}%
	}
	\caption{Average accuracy over all textures, averaged over exposure times, using the flow decode layer. Each texture accuracy includes a margin of error with a $95\%$ statistical confidence.}
	\label{tab:table_flowdecode_all}
\end{table*}

\begin{table*}
	\centering
	\resizebox{0.65\textwidth}{!}{%
	\csvreader[tabular=|l|l|,
	    table head=\hline Group & Short (300-600 ms.)\\\hline\hline,
	    late after line=\\\hline]%
	{tables/all_textures_avg_flowdecode_short.csv}{Name=\Name,Accuracy=\Accuracy}%
	{\Name & \Accuracy}%
	\csvreader[tabular=|l|,
	    table head=\hline Long (1200-4800 ms.)\\\hline\hline,
	    late after line=\\\hline]%
	{tables/all_textures_avg_flowdecode_long.csv}{Accuracy=\Accuracy}%
	{\Accuracy}%
	\csvreader[tabular=|l|,
	    table head=\hline All (300-4800 ms.)\\\hline\hline,
	    late after line=\\\hline]%
	{tables/all_textures_avg_flowdecode_all.csv}{Accuracy=\Accuracy}%
	{\Accuracy}%
	}
	\caption{Average accuracy over all textures, averaged over a range of exposure times, using the flow decode layer. Each texture accuracy includes a margin of error with a $95\%$ statistical confidence.}
	\label{tab:table_flowdecode_all_range}
\end{table*}
\clearpage
\clearpage
\twocolumn

\end{document}